\documentclass[lettersize,journal]{IEEEtran}
\usepackage{amsmath,amsfonts}
\usepackage{algorithmic}
\usepackage{algorithm}
\usepackage{array}
\usepackage[caption=false,font=normalsize,labelfont=sf,textfont=sf]{subfig}
\usepackage{textcomp}
\usepackage{stfloats}
\usepackage{url}
\usepackage{verbatim}
\usepackage{graphicx}
\usepackage[dvipsnames]{xcolor}
\usepackage[numbers]{natbib}
\usepackage{acro}
\usepackage{cancel}
\usepackage{subfig}
\usepackage{empheq}
\usepackage{xcolor}
\usepackage{graphicx}
\usepackage{adjustbox}
\usepackage{orcidlink}

\usepackage{titlesec}
\titlespacing*{\subsection}{0pt}{7pt}{0pt} 
\titlespacing*{\section}{0pt}{7pt}{0pt} 
\newcommand{\y}{\mathbf{y}}
\newcommand{\x}{\mathbf{x}}
\newcommand{\vb}{\mathbf{v}}
\newcommand{\vu}{\mathbf{u}}
\newcommand{\vnull}{\mathbf{0}} 
\newcommand{\vtheta}{\boldsymbol{\theta}}
\newcommand{\e}{\boldsymbol{\varepsilon}}
\newcommand{\A}{\mathbf{A}}
\newcommand{\I}{\mathbf{I}}
\newcommand{\Y}{\mathbf{Y}}
\newcommand{\Sig}{\mathbf{\Sigma}}

\DeclareMathOperator*{\argmax}{arg\,max}

\usepackage{bbm}

\usepackage{amsthm}
\newtheorem{theorem}{Theorem}[section]
\newtheorem{lemma}[theorem]{Lemma}

\newcommand{\norm}[1]{\left\lVert#1\right\rVert}

\DeclareAcronym{MLE}{short = MLE, long = maximum likelihood estimation}
\DeclareAcronym{EM}{short = EM, long = expectation-maximization}
\DeclareAcronym{IMS}{short = IMS, long = imaging mass spectrometry}
\DeclareAcronym{HSI}{short = HSI, long = hyperspectral imaging}
\DeclareAcronym{GMM}{short = GMM, long = Gaussian mixture model}
\DeclareAcronym{SMM}{short= SMM, long = spiked mixture model}
\DeclareAcronym{KMC}{short= {$k$MC}, long = {$k$-means clustering}}
\DeclareAcronym{SNR}{short= SNR, long = signal-to-noise ratio}
\DeclareAcronym{mz}{short= \emph{m/z}, long = mass-over-charge}
\DeclareAcronym{SIMO}{short= SIMO, long = single-input multiple-output}
\DeclareAcronym{MIMO}{short= MIMO, long = multiple-input and multiple-output}
\DeclareAcronym{LOD}{short=LOD, long = limit-of-detection}
\DeclareAcronym{QTOF}{short=QTOF, long = Quadrupole Time-of-Flight}
\DeclareAcronym{BIC}{short = BIC, long = Bayesian information criterion}
\DeclareAcronym{AIC}{short = AIC, long = Akaike information criterion}

\definecolor{manualcolor}{RGB}{255, 0, 0}
\definecolor{manualcolor}{RGB}{0, 0, 0}

\begin{document}

\title{Signal Recovery Using a Spiked Mixture Model}

\author{
    \IEEEauthorblockN{
        Paul-Louis Delacour\IEEEauthorrefmark{1},
        Sander Wahls\IEEEauthorrefmark{2}, \textit{Senior Member, IEEE}, 
        Jeffrey M. Spraggins\IEEEauthorrefmark{3}\IEEEauthorrefmark{5}\IEEEauthorrefmark{6},\\ \hspace{2.5cm}
        Lukasz Migas\IEEEauthorrefmark{1},
        and Raf Van de Plas\IEEEauthorrefmark{1}\IEEEauthorrefmark{3}, \textit{Member, IEEE}
    }\newline
        
\IEEEauthorblockA{
\IEEEauthorrefmark{1}Delft University of Technology, Delft Center for Systems and Control, Delft, Netherlands\\
\IEEEauthorrefmark{2}Karlsruhe Institute of Technology, Institute of Industrial Information Technology, Karlsruhe, Germany\\
\IEEEauthorrefmark{3}Vanderbilt University,  Dept. of Biochemistry \& Mass Spectrometry Research Center, Nashville, TN, U.S.A.\\
\IEEEauthorrefmark{5}Vanderbilt University, Dept. of Cell and Developmental Biology \& Dept. of Chemistry, TN, U.S.A.\\
\IEEEauthorrefmark{6}Vanderbilt University Medical Center, Dept. of Pathology, Microbiology, and Immunology, TN, U.S.A.\\
Corresponding authors: p.l.delacour@tudelft.nl \& raf.vandeplas@tudelft.nl
}
}

\markboth{Preprint, JUL 2025}%
{Delacour \MakeLowercase{\textit{et al.}}: Signal Recovery Using a Spiked Mixture Model}
\maketitle

\makeatletter
\newcommand{\manualtitle}{
    \twocolumn[
    \begin{center}
        {\LARGE \@title \\(Supplementary Material) \par}
        \vskip 1em
        {\normalsize \@author \par}
        \vskip 1em
    \end{center}
    ]
}
\makeatother
\begin{abstract}
We introduce the \acf{SMM} to address the problem of estimating a set of signals from many randomly scaled and noisy observations.
Subsequently, we design a novel \acf{EM} algorithm to recover all parameters of the \ac{SMM}.
Numerical experiments show that in low \acl{SNR} regimes, and for data types where the \ac{SMM} is relevant, \ac{SMM} surpasses the more traditional \acf{GMM} in terms of signal recovery performance.
The broad relevance of the \ac{SMM} and its corresponding \ac{EM} recovery algorithm is demonstrated by applying the technique to different data types.
The first case study is a biomedical research application, utilizing an \acl{IMS} dataset to explore the molecular content of a rat brain tissue section at micrometer scale.
The second case study demonstrates \ac{SMM} performance in a computer vision application, segmenting a \acl{HSI} dataset into underlying patterns.
While the measurement modalities differ substantially, in both case studies \ac{SMM} is shown to recover signals that were missed by traditional methods such as \acl{KMC} and \ac{GMM}.
\end{abstract}


\section{Introduction}
\noindent
Many advances in sensor technology and instrumentation are driven by the demand for greater specificity, sensitivity, and resolution, often implicitly leading to the acquisition of ever larger amounts of increasingly high-dimensional data.
This trend can be observed across a broad range of measurement modalities and application domains, including super-resolution imaging \cite{nobel2_Betzig2006}, novel sensor types for computer vision and remote sensing \cite{remote_Hagen2013ReviewOS}, chemical assays \cite{baeumner2022advancements}, communication \cite{grant_Free_massive}, and military applications \cite{artan2022future}.
In cases where limited signal strength or energy needs to be spread over a growing number of observations and dimensions, it can become increasingly difficult to differentiate between signals and to discern them from background variation.
For example, in certain imaging techniques, sampling at smaller spatial distances can degrade the \acf{SNR} of pixel-specific measurements, and adjusting the spectral resolution of a spectrometer can impact its \acf{LOD}.
Therefore, the ability to effectively and reliably recover signals from increasingly noisy measurements is becoming essential to unlocking the full potential of certain measurement modalities, particularly in scenarios with substantial background noise, low \ac{SNR}, or high \ac{LOD} (\textit{e.g.}, single-cell measurements with limited molecular material to measure).

Signal recovery in noisy environments without prior knowledge of the subpopulations within the observation pool is often conducted using a \acf{GMM} and by \textcolor{manualcolor}{\ac{MLE}} of the model's parameters through \acf{EM} optimization.
For example, motivated by multi-reference alignment in cryo-electron microscopy (a.k.a. the orbit retrieval problem), \citet{katsevich2023likelihood} studied likelihood maximization for the \ac{GMM} in low \ac{SNR} regimes.
The use of a standard \ac{GMM} implicitly assumes that the signal subpopulations or mixture components are normally distributed.
Although this is a broad assumption that fits many scenarios and contributes to the popularity of this approach, certain measurement types allow for more refined assumptions on the underlying mixture components. 
For those measurement types, using a \ac{GMM} will lead to suboptimal signal recovery, especially at low \ac{SNR}.

To address the mismatch between one such signal type and the \ac{GMM}, we introduce an alternative mixture model, called the \acf{SMM}.
In the \ac{SMM}, an observation or measurement is a randomly scaled version of one of a set of underlying signals called spikes, further perturbed by additive noise. 
Spiked models, introduced by \citet{johnstone2001distribution}, are a class of models characterized by the insertion of a planted vector into a random matrix. 
While previous work explored the statistical properties of these models under various prior distributions on the spike \cite{perry2018optimality}, our \ac{SMM} takes a distinct approach. 
One can view the \ac{SMM}'s covariance matrix as a sum of spiked Wishart matrices with one degree of freedom and each spike sampled from a categorical prior.
This differs from the multi-spiked tensor model in \cite{multi_spike_ben_arous} as we focus on a mixture model rather than on linear combinations of spikes.

Although the \ac{SMM}'s signal spikes could potentially be recovered using a \ac{GMM}, we show that \ac{GMM}-based recovery is only practical in high-\ac{SNR} scenarios.
In contrast, an \ac{EM}-based approach that directly estimates an \ac{SMM} consistently outperforms the \ac{GMM}, especially in noisy conditions.
While one could argue that refining the Gaussian distribution assumption makes the \ac{SMM} less broadly applicable than the \ac{GMM}, the \ac{SMM} remains widely relevant to a variety of application domains.
In those domains, the \ac{SMM} tends to fit the underlying signal model better, enabling advanced signal detection and recovery.
This becomes especially valuable when addressing high-noise, low-\ac{SNR} measurements.
Beyond its primary function of recovering signals from noisy measurements, an \ac{EM}-driven approach for \ac{SMM}-fitting provides additional information that classical methods, such as \acf{KMC}, do not.
These secondary outputs from the fitting process include an implicit estimate of the observations' noise variance, mixture probabilities for each spike, and spike responsibilities, \textit{i.e.}, the probability that a noisy observation belongs to a specific spike.

The desirable asymptotic properties of \ac{MLE} make it a common choice to drive the fitting process.
As the number of observations goes to infinity, \ac{MLE} is an asymptotically consistent and efficient estimator, \textit{i.e.}, it converges to the true parameter values and achieves the lowest possible variance among unbiased estimators (\cite{casella2024statistical}, Chap. 10).
However, computing the \ac{MLE} remains a challenge in many scenarios.
A typical algorithmic procedure to find a \ac{MLE} candidate in the presence of unobserved latent variables is \ac{EM} \cite{gamp_based_Low_Complexity,inverse_EM,mixtures_elliptical_dist}.
\Ac{EM} is a sequential algorithm performing `soft assignment' of observations to mixture components, with guarantees to converge to a local maximum.
Since deriving the equations for an \ac{EM} optimization is model-specific, Section~\ref{sec:expected_maximization} is dedicated to developing a custom \ac{EM} algorithm for the \ac{SMM}.

In Section~\ref{sec:gmm_comparison}, we provide a direct comparison between standard \ac{GMM}-based signal recovery and the proposed \ac{SMM}. Using a synthetic dataset with known ground truth signal populations (spikes), we demonstrate that in low-noise scenarios \ac{SMM} and \ac{GMM} achieve equivalent recovery. However, in high-noise regimes, \ac{SMM} substantially outperforms \ac{GMM}. 

Section~\ref{sec:applications} demonstrates the \ac{SMM} in real-world applications.
First, we use the \ac{SMM} \ac{EM}-algorithm to recover underlying molecular signatures from noisy \acf{IMS} measurements of a rat brain tissue section.
The recovered spikes align with known biological structures, the estimated responsibilities segment the tissue according to molecular content, and \ac{SMM} retrieves histological patterns missed by \ac{GMM}.
This case illustrates \ac{SMM}'s enhanced signal accuracy when recovering in the presence of sizeable noise.
The second application uses \ac{SMM} to segment \acf{HSI} measurements.
This case study demonstrates \ac{SMM}'s ability to differentiate signal subpopulations that \ac{GMM} and \ac{KMC} have difficulty with.
Notably, the \ac{SMM} is not intrinsically related to imaging data. 
Its signal model also finds relevance in domains such as wireless communication, where the \ac{SMM} describes a random access narrowband flat fading \acl{SIMO} communication system (\cite{tsefundamentals}, Sec. 7.3).
Imaging examples allow \ac{SMM} estimation results (\textit{e.g.}, responsibilities) to be shown as images, aiding interpretation.

\IEEEpubidadjcol

\section{Expectation-Maximization algorithm for the spiked mixture model}
\label{sec:expected_maximization}
\begin{figure}[t]
    \centering
    \includegraphics[width=3.5in]{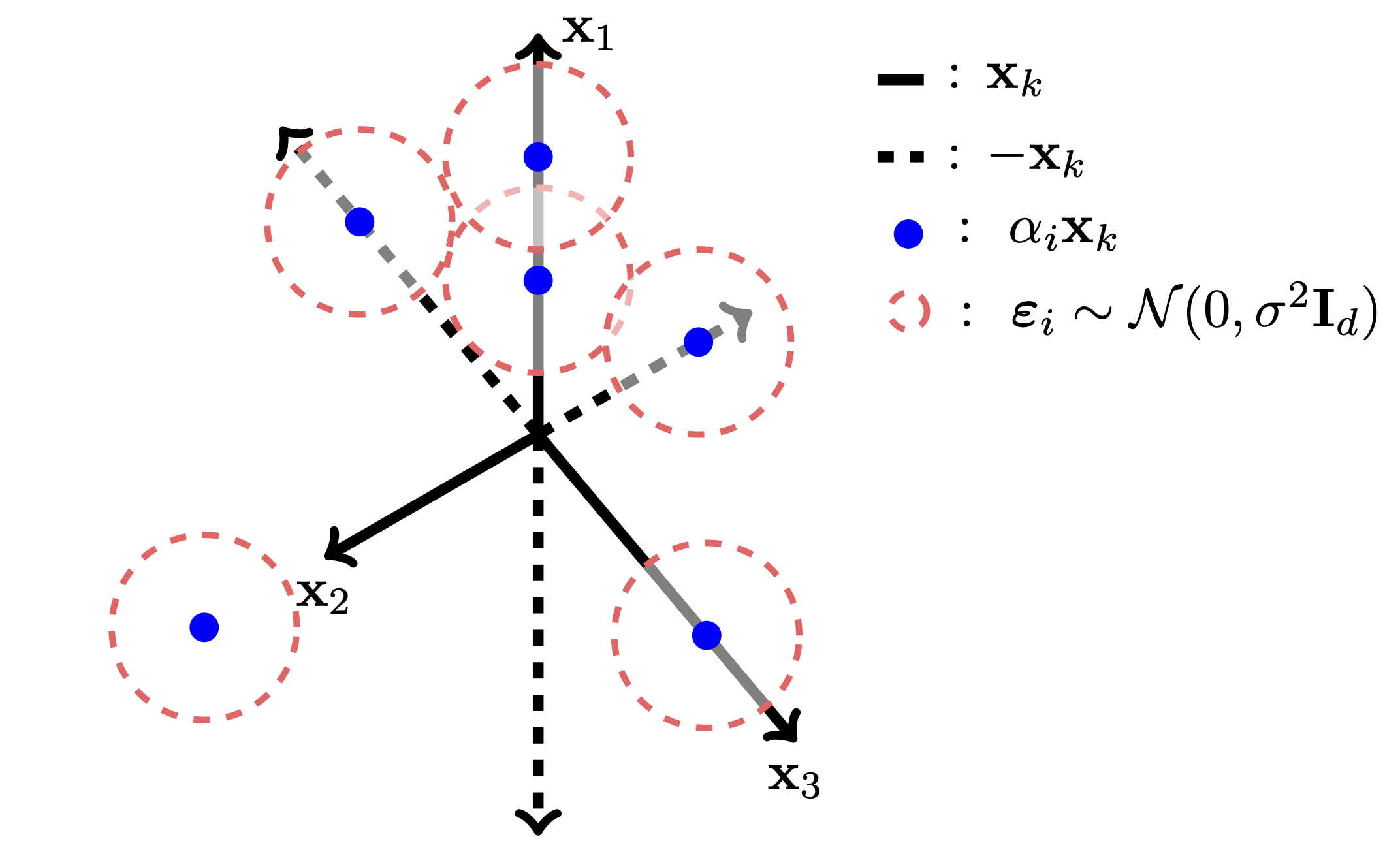}
    \caption{Examplary realization of a spiked mixture model as specified in \eqref{eq:model} that underlies $N$ observations, with $N=6$, $d=2$, and $K=3$. Black lines represent the directions, or spikes, along which observations concentrate, blue dots are scaled observations along a particular spike without noise, and red circles represent one standard deviation of the Gaussian noise perturbations.}
    \label{fig:multi-spiked}
\end{figure}

\noindent
We study the problem of estimating a set of signals from observations that consist of randomly scaled and noisy copies of those signals. More precisely, we consider $N$ independent observations $\y_1, \ldots, \y_N \in \mathbb{R}^d$, sampled from the model
\begin{align}
     & \y = 
    \begin{cases}
        \alpha \x_1 + \e &\text{with probability } \pi_1\\
        \quad\vdots \\
        \alpha \x_{K} + \e &\text{with probability } \pi_{K}\\
    \end{cases},\label{eq:model} \\
    &\alpha \sim \mathcal{N}(0,1),\: \e \sim \mathcal{N}(\vnull,\sigma^2\I), \nonumber\\
    &\sum_{k=1}^K \pi_k = 1, \: \x_1, \ldots, \x_K \in \mathbb{R}^d  \nonumber,
\end{align}
where $\alpha$ is the random scaling factor of observation $\y$, $\x_k$ is the $k$-th subpopulation or spike, $\e$ is the random noise of observation $\y$, and $\pi_k$ is the probability of the $k$-th subpopulation.
Let $z\in\{1,\dots,K\} $ be a latent categorical variable indicating which of the spikes $\x_1, \ldots , \x_K$ was used to generate $\y$.
Given $z$, the random vector $\y$  is normally distributed as it is a sum of independent normally distributed variables.
Specifically, by computing $\mathbb{E}[\alpha \x_z + \e]$ and $\mathbb{E}[(\alpha \x_z + \e)(\alpha \x_z +\e)^\mathrm{T}]$, we find that $ \y|z \sim \mathcal{N}(\vnull,\boldsymbol{\Sigma}_z : =\x_z \x_z^\mathrm{T} + \sigma^2 \textbf{I})$. 
Model~\eqref{eq:model} is thus a constrained \acf{GMM} with density
\begin{align}
    p(\y) = \sum_{k=1}^{K} \pi_k \: p_{\mathcal{N}(\vnull, \boldsymbol{\Sigma}_k)}(\y),
    \label{eq:p-of-y}
\end{align}
where $p_{\mathcal{N}(\vnull, \boldsymbol{\Sigma}_z)}$ denotes the probability density function of a multivariate Gaussian distribution with mean $\vnull$ and covariance $\boldsymbol{\Sigma}_z$.
Instead of recovering the mean and covariance matrix of each Gaussian component, our goal here is to estimate the vector of parameters $\vtheta = \{\x_1 , \ldots , \x_K, \pi_1, \ldots , \pi_K , \sigma^2 \}$ that defines them.   
We refer to this model as a \emph{spiked mixture} model.
\begin{itemize}
    \item The adjective \emph{spiked} refers to observations concentrating along certain directions, called spikes, in the measurement space (Figure~\ref{fig:multi-spiked} shows an exemplary realization).
    \item \emph{Mixture} refers to there being a set of $K$ directions, spikes, or signal subpopulations within the observations.
\end{itemize}

\noindent
As $\y|z$ is normally distributed, using Sylvester's determinant identity and the Sherman-Morrison formula to compute the determinant and the inverse of the covariance matrix $\boldsymbol{\Sigma}_z := \x_z \x_z^\mathrm{T} + \sigma^2 \I$, we find the conditional density
\begin{align}
\begin{split}
    p_{\vtheta}(\y|z) 
    &= \frac{1}{\sqrt{(2 \pi)^d}}\exp\left[\frac{-1}{2\sigma^2}\left(\|\y\|^2 - \frac{(\y^\mathrm{T}\x_z)^2}{ \|\x_z\|^2+ \sigma^2}\right) \right.\\
    &\quad \left. -\frac{1}{2} \ln(\|\x_z\|^2 + \sigma^2) - (d-1) \ln \sigma \right],  \label{eq:likelihood}
\end{split}
\end{align}
where the subscript $\vtheta$ indicates that a density is determined by the parameter vector.
With $N$ observations $\y_1, \ldots, \y_N$ from \eqref{eq:model}, the log-likelihood corresponds to
\begin{align*}
\begin{split}
\ln p_{\vtheta}(\y_1, \ldots , \y_N )
 &= \sum_{i=1}^N \ln \sum_{k=1}^K \pi_k \exp\left[\frac{1}{2\sigma^2}\frac{(\y_i^\mathrm{T} \x_k)^2}{\|\x_k\|^2+\sigma^2}\right. \\
 &\quad \left. -\frac{1}{2}\ln(\|\x_k\|^2 + \sigma^2)- (d-1)\ln \sigma \right] + C,
\end{split}
\end{align*}
where we use $p_{\vtheta}(\y_1, \ldots , \y_N ) = \prod_{i=1}^N p_{\vtheta}(\y_i)$ and exploit \eqref{eq:p-of-y} and \eqref{eq:likelihood}. The constant $C$ is independent of $\vtheta$.
\textcolor{manualcolor}{Note that one can only hope to recover the vectors $\x_1, \ldots, \x_K$ in the model (1) up to a sign change since it suffers from an intrinsic symmetry. Technically, it is thus non-identifiable \cite{holzmann2006identifiability}.}
Indeed, for two sets of parameters, $\vtheta_1$ and $\vtheta_2$, that are the same up to a sign change in front of the $\x_k$s we have: 
\begin{align*}
    p_{\vtheta_1}(\y_1,\ldots,\y_N) &= p_{\vtheta_2}(\y_1,\ldots,\y_N).
\end{align*}

\noindent
The \acf{EM} algorithm \cite{dempster1977maximum}, originally developed by \citeauthor{dempster1977maximum} in \citeyear{dempster1977maximum}, is a common approach to finding a candidate maximum likelihood estimate of $\vtheta$.
It is an iterative procedure that is made up of two steps.
Let $\vtheta^{[t]} = \left\{\hat{\x}_1 , \ldots , \hat{\x}_{K}, \hat{\pi}_1, \ldots , \hat\pi_{K}, \hat{\sigma}^2\right\}$ be the estimate of $\vtheta$ at step $t$.
The first step, called the expectation or $E$-step, computes
\begin{align*}
    \rho_z^{[i]} &:= p_{\vtheta^{[t]}}(z | \y_i) &\forall i \in \{1,\ldots , N\}, \ z \in \{1, \ldots , K\}.
\end{align*}
Using Bayes rule, we have
\begin{align*}
    p_{\vtheta^{[t]}}(z | \y_i) = \frac{p_{\vtheta^{[t]}}(\y_i | z)\pi_z}{\sum_{k = 1}^K p_{\vtheta^{[t]}}(\y_i | z=k) \pi_k} .
\end{align*}
Using \eqref{eq:likelihood} and simplifying the terms that cancel in the numerator and denominator, we get 
\begin{align*}
    \rho_z^{[i]} := p_{\vtheta^{[t]}}(z | \y_i) = \frac{\Tilde{\rho}_z^{[i]}}{\sum_{k=1}^{K} \Tilde{\rho}_k^{[i]}}, \quad \textrm{where}
\end{align*}
\begin{align*}
    \Tilde{\rho}_z^{[i]} := \hat{\pi}_z  \exp\left(\frac{1}{2\hat{\sigma}^2} \frac{(\y_i^\mathrm{T} \hat{\x}_z)^2}{\|\hat{\x}_z\|^2+\hat{\sigma}^2} -\frac{1}{2}\ln(\|\hat{\x}_z\|^2+\hat{\sigma}^2)\right).
\end{align*}

\noindent
The second step, called the maximization or $M$-step, computes $\vtheta^{[t+1]}$ by finding the feasible $\vtheta$ maximizing 
\begin{align}
\begin{split}
    &\mathcal{Q}(\vtheta; \vtheta^{[t]}) \\
    &= \sum_{i=1}^N \mathbb{E}_{z\sim p_{\vtheta^{[t]}}(\cdot|\y_i)} \left[ \ln p_{\vtheta}(\y_i ,z)\right]\\
    &= \sum_{i=1}^N \sum_{k=1}^K \rho_{k}^{[i]} \left[ \ln(\pi_k) -\frac{1}{2\sigma^2}\left(\|\y_i\|^2 - \frac{(\y_i^\mathrm{T} \x_k)^2}{\|\x_k\|^2 + \sigma^2}\right) \right.\\
    &\quad\left. -\frac{1}{2} \ln(\|\x_k\|^2 + \sigma^2) - \frac{d}{2} \ln(2\pi) -(d-1)\ln(\sigma)
    \right].
\end{split}
\label{eq:Q_function_SMM}
      \end{align}

\noindent
Maximizing the expectation function $\mathcal{Q}$ acts as a proxy for maximizing the log-likelihood and guarantees the following improvement (see (3.10) in \cite{dempster1977maximum}):
\begin{align*}
    &\ln p_{\vtheta}(y_1,\ldots,y_n) - \ln p_{\vtheta^{[t]}}(y_1,\ldots,y_n) \\
    &\geq \mathcal{Q}(\vtheta;\vtheta^{[t]})-\mathcal{Q}(\vtheta^{[t]};\vtheta^{[t]}).
\end{align*}
In other words, at every step, the log-likelihood improves by at least as much as the $\mathcal{Q}$ function.
Precisely, the $M$-step is
\begin{equation}
    \begin{array}{lll}
         \vtheta^{[t+1]} =& \argmax & \mathcal{Q}(\vtheta; \vtheta^{[t]}) \\
         &\vtheta \quad\textrm{s.t} 
         & \x_k \in \mathbb{R}^d \\
         && \pi_k \geq 0\\
         && \sum_{k=1}^{K} \pi_k = 1\\
         && \sigma \in \mathbb{R}_+ 
    \end{array}.
\end{equation}

\noindent
To solve this, we instead look at the less constrained problem
\begin{equation}
    \begin{array}{lll}
         \vtheta^{[t+1]} =& \argmax & \mathcal{Q}(\vtheta; \vtheta^{[t]}) \\
         &\vtheta \quad\text{s.t} 
         & \sum_{k=1}^{K} \pi_k = 1
    \end{array},
\end{equation}
and verify that a found $\vtheta^{[t+1]}$ satisfies the missing constraints.
The solution of this less constrained $M$-step is, by the Karush-Kuhn-Tucker conditions, a critical point of the Lagrangian
\begin{align*}
    \mathcal{L} &= \mathcal{Q}(\vtheta; \vtheta^{[t]}) + \lambda \left(1 - \sum_{k=1}^{K} \pi_k\right).
\end{align*}
In what follows, we find a critical point of this Lagrangian: 
\begin{align}
    \left(\frac{\partial \mathcal{L}}{\partial \x_1}, \ldots , \frac{\partial \mathcal{L}}{\partial \x_K} , \frac{\partial\mathcal{L}}{\partial \pi_1}, \ldots , \frac{\partial \mathcal{L}}{\partial\pi_K}, \frac{\partial\mathcal{L}}{\partial\sigma},\frac{\partial\mathcal{L}}{\partial\lambda}\right) = \vnull.
    \label{eq:critical-point-lagrangian}
\end{align}
In order to simplify the notation for $Q(\vtheta;\vtheta^{[t]})$, and thus $\mathcal{L}$, we define the following for all $e \in \{1,\ldots, K\}$:
\begin{align}
    \A_e &:= \sum_{i=1}^{N} \rho_e^{[i]} \y_i \y_i^\mathrm{T} \label{eq:Ae} &&\in \mathbb{R}^{d\times d}, \\
    \gamma_e &:= \sum_{i=1}^N \rho_e^{[i]} \label{eq:rho_e} &&\in \mathbb{R}, \\
    \Y &:= \begin{bmatrix}
    \vert & & \vert \\
    \y_1   & ... & \y_N   \\
    \vert & & \vert 
\end{bmatrix} &&\in \mathbb{R}^{d\times N}.
\end{align}
With this and the fact that $\sum_{k=1}^K \rho_{k}^{[i]}=1$ for all $i$, we can rewrite $\mathcal{Q}(\vtheta; \vtheta^{[t]})$ as
\begin{align}
    \mathcal{Q}(\vtheta; \vtheta^{[t]}) &= \sum_{k=1}^K \gamma_k \ln \pi_k -\frac{\|\Y\|_\mathrm{F}^2}{2\sigma^2} + \frac{1}{2\sigma^2} \sum_{k=1}^K\frac{\x_k^\mathrm{T} \A_k \x_k}{\|\x_k\|^2 + \sigma^2} \nonumber \\
    & -\frac{1}{2} \sum_{k=1}^K \gamma_k \ln(\|\x_k\|^2+\sigma^2) -(d-1)N \ln(\sigma) \nonumber\\
    &-\frac{dN}{2} \ln(2\pi).
    \label{eq:Q_function}
\end{align}

\setlength{\parindent}{0pt}
\paragraph{Derivatives with respect to $\mathbf{\pi}_e$ for $e\in \{1,\ldots , K\}$}

The condition in \eqref{eq:critical-point-lagrangian} is reached when
\begin{align*}
    \frac{\partial \mathcal{L}}{\partial \pi_e} &= \frac{\gamma_e}{\pi_e} - \lambda = 0,
\end{align*}
or, equivalently, when
\begin{align}
    \lambda &= \frac{1}{\pi_e} \gamma_e &\forall e\in \{1,\ldots, K\}.
    \label{eq:lambda_lagrange}
\end{align}
Using~\eqref{eq:lambda_lagrange} and the constraint $\sum_{k=1}^K \pi_k =1$, which follows from the condition $\frac{d\mathcal{L}}{d\lambda}=0$ in \eqref{eq:critical-point-lagrangian}, we find
\begin{align*}
    1 = \sum_{k=1}^K \pi_k = \sum_{k=1}^K \frac{\gamma_k}{\lambda} = \frac{N}{\lambda},
\end{align*}
which gives $\lambda = N$, and  
\begin{align}
    \pi_e &= \frac{\gamma_e}{N} &\forall e\in \{1,\ldots , K\}. \label{eq:pi}
\end{align}
We note here that $\pi_e$ from~\eqref{eq:pi} is indeed in the range $[0,1]$.

\paragraph{Derivatives with respect to $\mathbf{x}_e$ for $e\in \{1,\ldots , K\}$}
We now expand the first condition in \eqref{eq:critical-point-lagrangian}. The derivative of the Lagrangian w.r.t. $\x_e$ is
\begin{align*}
    \frac{\partial \mathcal{L}}{\partial \x_e} &= \frac{1}{2\sigma^2}\left[\frac{2(\|\x_e\|^2+\sigma^2)\A_e \x_e - 2 (\x_e^\mathrm{T} \A_e \x_e) \x_e}{(\|\x_e\|^2+\sigma^2)^2}\right] \\
    &\quad- \frac{\gamma_e \x_e}{\|\x_e\|^2 + \sigma^2}.
\end{align*}
Setting it to zero and rearranging the terms, we get 
\begin{align}
  \A_e \x_e &= \left(\frac{\x_e^\mathrm{T} \A_e \x_e}{\|\x_e\|^2+\sigma^2} + \sigma^2 \gamma_e \right)  \x_e .
   \label{eq:critic_xe}  
\end{align}
We distinguish two possible cases:
\begin{itemize}
     \item either $\x_e$ is the zero vector, and this is trivially satisfied, 
     \item or $\x_e$ is different from the zero vector.
\end{itemize}
In the latter case, \eqref{eq:critic_xe} means that $\x_e$ is an eigenvector of $\A_e$ with eigenvalue $\lambda_e$:
     \begin{align}
         \A_e \x_e &=: \lambda_e \x_e.
         \label{eq:eigenvalue_problem}
     \end{align}
Projecting \eqref{eq:critic_xe} onto $\x_e/\|\x_e\|^2$, we find
     \begin{align*}
         \frac{\x_e^\mathrm{T} \A_e \x_e}{\|\x_e\|^2} &= \frac{\x_e^\mathrm{T} \A_e \x_e}{\|\x_e\|^2 + \sigma^2} + \sigma^2 \gamma_e.
     \end{align*}
Substituting \eqref{eq:eigenvalue_problem} into this equation, we get
     \begin{align*}
         \lambda_e &= \lambda_e \frac{\|\x_e\|^2}{\|\x_e\|^2+\sigma^2} + \sigma^2 \gamma_e,
     \end{align*}
which simplifies to
     \begin{align*}
         \lambda_e &= (\|\x_e\|^2 + \sigma^2)\gamma_e &\forall e\in \{1,\ldots , K\}.
     \end{align*}
Thus, when $\x_e \neq \vnull$, the critical point in \eqref{eq:critical-point-lagrangian} satisfies
     \begin{align}
       \|\x_e\|^2 &=  \frac{\lambda_e}{\gamma_e}-\sigma^2  &\forall e\in \{1,\ldots , K\}. \label{eq:lambda}
     \end{align}

\paragraph{Derivative with respect to $\sigma$}

Finally, we expand the third condition in \eqref{eq:critical-point-lagrangian}. The corresponding derivative is
\begin{align*}
\begin{split}
    \frac{\partial \mathcal{L}}{\partial \sigma} &= \frac{1}{\sigma^3}\|\Y\|_\mathrm{F}^2 -\frac{1}{\sigma^3}\sum_{k=1}^K \frac{\x_k^\mathrm{T} \A_k \x_k}{\|\x_k\|^2+\sigma^2} - \frac{1}{\sigma}\sum_{k=1}^K \frac{\x_k^\mathrm{T} \A_k \x_k}{(\|\x_k\|^2 + \sigma^2)^2} \\
    &\quad - \sigma \sum_{k=1}^K \frac{\gamma_k}{\|\x_k\|^2+\sigma^2} - \frac{(d-1)N}{\sigma} .
\end{split}
\end{align*}
Setting the prior equation to zero and multiplying by $\sigma^3$, yields 
\begin{align*}
    &\sigma^2 (d-1)N\\
    &= \|\Y\|_\mathrm{F}^2 -\sum_{k=1}^K \frac{\x_k^\mathrm{T} \A_k \x_k}{\|\x_k\|^2+\sigma^2} \\
    &\quad-\sigma^2 \sum_{k=1}^K \frac{\x_k^\mathrm{T} \A_k \x_k}{(\|\x_k\|^2 + \sigma^2)^2} 
    - \sigma ^4\sum_{k=1}^K \frac{\gamma_k}{\|\x_k\|^2+\sigma^2}\\
    &= \|\Y\|_\mathrm{F}^2 - \sum_{k : \x_k \neq \vnull} \frac{\x_k^\mathrm{T} \A_k \x_k}{\|\x_k\|^2+\sigma^2} - \sigma^2\sum_{k: \x_k = \vnull} \gamma_k \\
    &\quad - \sigma^2 \sum_{k : \x_k \neq \vnull}  \frac{1}{\|\x_k\|^2 + \sigma^2} \left(\frac{\x_k^\mathrm{T} \A_k \x_k}{\|\x_k\|^2+\sigma^2} + \sigma^2 \gamma_k \right). \\
\end{align*}
By rearranging the $\sigma^2$ terms, replacing $\gamma_k$ in the last term with \eqref{eq:lambda}, and using the eigenrelation \eqref{eq:eigenvalue_problem} twice, we get
\begin{align*}
 &\sigma^2 \left[(d-1)N+\sum_{k: \x_k = \vnull} \gamma_k\right]\\   
 &= \|\Y\|_\mathrm{F}^2 - \sum_{k:\x_k \neq \vnull} \left(\frac{\x_k^\mathrm{T} \A_k \x_k}{\|\x_k\|^2 + \sigma^2} + \sigma^2 \frac{\lambda_k}{\|\x_k\|^2 + \sigma^2}\right) \\
 &= \|\Y\|_\mathrm{F}^2 - \sum_{k:\x_k \neq \vnull} \left(\lambda_k \frac{\|\x_k\|^2 + \sigma^2}{\|\x_k\|^2 + \sigma^2}\right) \quad\textrm{by using \eqref{eq:eigenvalue_problem}.}
\end{align*}
This gives a final expression for $\sigma^2$: 
\begin{align*}
    \sigma^2 = \frac{ \|\Y\|_\mathrm{F}^2 -\sum_{k : \x_k \neq \vnull} \lambda_k}{(d-1)N + \sum_{k: \x_k = \vnull} \gamma_k}.
\end{align*}
Since $\sum_{k=1}^K \gamma_k = N$ implies
$\sum_{k:\x_k\neq \vnull} \gamma_k = N - \sum_{k:\x_k=\vnull} \gamma_k$, we can rewrite the expression for $\sigma^2$ as
\begin{align}
    \sigma^2 = \frac{ \|\Y\|_\mathrm{F}^2 -\sum_{k : \x_k \neq \vnull} \lambda_k}{dN - \sum_{k: \x_k \neq \vnull} \gamma_k}.
    \label{eq:sigma}
\end{align}
In Lemma~\ref{lem:sigma_positive} of the Appendix, we show that estimate \eqref{eq:sigma} of $\sigma^2$ is indeed positive. \\

\noindent
In sum, the equations for critical points, $\forall k \in \{1,\ldots K\}$, are:
\begin{align*}
    \pi_k &= \frac{\gamma_k}{N} &\textrm{from }\eqref{eq:pi} \\ 
    &\left\{
    \begin{array}{cc}
   \textrm{either } \x_k = \vnull & \\ 
   \textrm{or } \A_k \x_k= \lambda_k \x_k, &\|\x_k\|^2 = \frac{\lambda_k}{\gamma_k} - \sigma^2 
   \end{array}\right. &\textrm{from } \eqref{eq:eigenvalue_problem}, \eqref{eq:lambda}\\
   S &= \{k \in [K] \textrm{ s.t } \x_k \neq \vnull\}\\
   \sigma^2 &= \sigma^2(S):= \frac{ \|\Y\|_\mathrm{F}^2 -\sum_{k \in S} \lambda_k}{dN - \sum_{k \in S}\gamma_k }&\textrm{from } \eqref{eq:sigma}
\end{align*}

\noindent
Now that we have established the equations that characterize critical points, we will analyze which critical point maximizes $\mathcal{Q}(\vtheta,\vtheta^{[t]})$.
After a series of algebraic reformulations provided in Appendix~\ref{sec:app_Q_saddle_point}, we find that for a $\vtheta$ satisfying the equations of a critical point, the expression of $\mathcal{Q}$ simplifies to
\begin{align*}
    \mathcal{Q}(\vtheta;\vtheta^{[t]}) 
    &= \hat{C} -\frac{1}{2} \left[dN - \sum_{k \in S} \gamma_k \right]\ln \left(\frac{\|\Y\|_\mathrm{F}^2-\sum_{k\in S}\lambda_k}{dN - \sum_{k\in S} \gamma_k}\right) \\
    &-\frac{1}{2} \sum_{k \in S} \gamma_k \ln\left(\frac{\lambda_k}{\gamma_k}\right),
\end{align*}
with $\hat{C}$ a constant independent of $\vtheta$.
Taking the derivative of the previous expression with respect to $\lambda_k$ for $k\in S$ we see 
\begin{align}
    \frac{\partial \mathcal{Q}(\vtheta;\vtheta^{[t]})}{\partial \lambda_k} \geq 0 \iff\sigma^2(S) \leq \frac{\lambda_k}{\gamma_k} .
    \label{eq:equivalence_lambda}
\end{align}
For a $\lambda_j(\A_k)$ satisfying \eqref{eq:equivalence_lambda}, we know that any $\lambda_k > \lambda_j(\A_k)$ also satisfies \eqref{eq:equivalence_lambda} since it decreases $\sigma^2(S)$ and it increases $\frac{\lambda_k}{\gamma_k}$. 
From the equivalency in \eqref{eq:equivalence_lambda}, this means that $\mathcal{Q}(\vtheta;\vtheta^{[t]})$ is increasing on the set $[\lambda_j(\A_k),\lambda_1(\A_k)]$. As a result, picking $\lambda_k = \lambda_1(\A_k)$ for all $k$ maximizes $\mathcal{Q}$.

\noindent
We are now left with picking a set $S$ that maximizes the function $\mathcal{Q}(\vtheta;\vtheta^{[t]})$ or, equivalently, that minimizes the function
\begin{align*}
    g(S) = \left[dN - \sum_{k\in S} \gamma_k\right] \ln \sigma^2(S) + \sum_{k\in S} \gamma_k \ln \left(\frac{\lambda_k}{\gamma_k}\right).
\end{align*}
We note that not all sets $S$ satisfy the critical point equations. Indeed, for $k\in S$, the associated eigenvalue $\lambda_k$ must satisfy $\frac{\lambda_k}{\gamma_k}-\sigma^2(S) \geq 0$.
Therefore, we call $S \subseteq[K]$ a valid set when
\begin{align*}
    \sigma^2(S) &\leq \frac{\lambda_k}{\gamma_k} &\forall k \in S.
\end{align*}
We will denote by $\mathcal{V}$ the set of all valid sets: 
\begin{align*}
    \mathcal{V} = \left\{ S\subseteq [K] \mid \forall k \in S : \sigma^2(S) \leq \lambda_k/ \gamma_k \right\}.
\end{align*}
Note that $\mathcal{V}$ is never empty since it contains at least the empty set for which the condition is trivially satisfied.
The optimal set $S^\star$ is given by 
\begin{align*}
    \begin{array}{lll}
         S^\star = & \text{argmin} & g(S) \\ 
         & S \in \mathcal{V}
    \end{array}.
\end{align*}
A naive procedure trying all possible valid sets would take $\mathcal{O}(K2^K)$.
In Appendix~\ref{sec:appendix_greedy_opt}, we provide a $\mathcal{O}(K^2)$ procedure.\\

\noindent
Algorithm~\ref{alg:em} summarizes all operations performed by the \ac{EM} optimization to fit a $K$-\acl{SMM}.
\textcolor{manualcolor}{
Standard convergence results for the \ac{EM} algorithm can be applied to our algorithm. For details, see section~\ref{sec:convergence_results_em} of the supplementary material. 
}
\begin{algorithm}
\caption{EM fitting of a $K$-\acl{SMM}}\label{alg:em}
\begin{algorithmic}
\STATE
\STATE {\textsc{Initialization :}}
\STATE \hspace{0.5cm} $\x_1, \ldots , \x_K, \pi_1, \ldots , \pi_K , \sigma^2$
\STATE
\STATE {\textsc{Repeat until convergence:}}
\STATE \hspace{0.2cm} $\textbf{E-Step:} $
\STATE \hspace{0.4cm} \textbf{for} $ i = 1,\ldots,N, e = 1, \ldots , K $
\STATE \hspace{0.8cm} $\Tilde{\rho}_e^{[i]} 
    = \pi_e \cdot \exp\left[\frac{1}{2 \sigma^2} \frac{(\y_i^\mathrm{T} \x_e)^2}{\|\x_e\|^2+\sigma^2} - \frac{1}{2} \ln (\|\x_e\|^2 + \sigma^2)\right] $ 
\STATE \hspace{0.8cm} $ \rho_e^{[i]} = \frac{\Tilde{\rho}_e^{[i]}}{\sum_{k=1}^{K} \Tilde{\rho}_k^{[i]}}  $
\STATE
\STATE \hspace{0.2cm} $\textbf{M-Step: }$ 
\STATE \hspace{0.4cm} \textbf{for} $e = 1,...,K $ :
\STATE \hspace{0.8cm}$ \gamma_e = \sum_{i=1}^{N} \rho_e^{[i]}$
\STATE \hspace{0.8cm} $\pi_e = \gamma_e/N$
\STATE \hspace{0.8cm} $\A_e = \sum_{i=1}^N \rho_e^{[i]} \y_i \y_i^\mathrm{T}$
\STATE \hspace{0.8cm} $\lambda_e , \vb_e \text{ largest eigenvalue/vector pair of }\mathbf{A}_e $
\STATE \hspace{0.4cm} \textbf{Find optimal} S:
\STATE \hspace {0.8cm}  $\sigma^2(S) = \left[ \|\Y\|_\mathrm{F}^2 - \sum_{k\in S} \lambda_k \right]/(dN - \sum_{k \in S} \gamma_k) $
\STATE \hspace{0.8cm}  $\mathcal{V} = \left\{ S\subseteq [K] \:\text{s.t.}\:\: \forall k \in S : \sigma^2(S) \leq \lambda_k/ \gamma_k \right\}$
\STATE \hspace{0.8cm} $g(S) = \left[dN - \sum_{k\in S} \gamma_k\right] \ln \sigma^2(S) + \sum_{k\in S}\gamma_k \ln \frac{\lambda_k}{\gamma_k}$
\STATE \hspace{0.8cm} $S^\star = \text{argmin}_{S \in \mathcal{V}} \:\:g(S)$
\STATE \hspace{0.8cm} $\sigma^2 = \sigma^2(S^\star)$
\STATE \hspace{0.4cm} \textbf{for} $e = 1,...,K $ :
\STATE \hspace{0.8cm} \textbf{if} $e \in S^\star$ :
\STATE \hspace{1.2cm} $\x_e = \sqrt{\frac{\lambda_e}{\gamma_e} - \sigma^2} \cdot  \vb_e$
\STATE \hspace{0.8cm} \textbf{else}:
\STATE \hspace{1.2cm} $\x_e = 0$
\end{algorithmic}
\end{algorithm}

\section{Comparison with standard \ac{GMM}}
\label{sec:gmm_comparison}
\noindent
We now compare our \ac{SMM} method to a \ac{GMM}-based approach.
Since our model is a constrained \acl{GMM}, we can use a standard \ac{GMM} to estimate the covariance matrices, yielding $\hat{\Sigma}_{1}, \ldots, \hat{\Sigma}_K$.
If this recovery is successful, we expect the following relationships to hold (albeit up to a possible relabeling of the $\x_k$s): 
\begin{align*}
    \hat{\Sigma}_k &= \x_k \x_k^\mathrm{T} + \sigma^2I &\forall k \in [K].
\end{align*}
Once the covariances are estimated, we can extract the $\x_k$s by solving the following optimization problem:
\begin{equation}
    \begin{array}{lcl}
        &\min & \sum_{k=1}^K \|\hat{\Sigma}_k - \left(\hat{\x}_k\hat{\x}_k^\mathrm{T}+ \hat{\sigma}^2 I\right)\|_\mathrm{F}^2 \\
    &\hat{\x}_1,\ldots,\hat{\x}_K,\hat{\sigma}^2
    \end{array}
    \label{eq:optim_GMM}.
\end{equation}
Let $\lambda_i(\Sigma_k)$ denote the $i$-th largest eigenvalue of $\Sigma_k$ and $\vb_i(\Sigma_k)$ its associated eigenvector. 
Lemma~\ref{lem:minimizer_gmm} in appendix~\ref{app:gmm_proof} shows that whenever the condition 
\begin{align}
    \lambda_1(\hat{\Sigma}_j) &\geq \frac{\sum_{k=1}^K \sum_{i=2}^d \lambda_i(\hat{\Sigma}_k)}{K(d-1)} &\forall j \in [K] \label{eq:larg_spectral_gap}
\end{align}
holds, the solution to 
\eqref{eq:optim_GMM} is given by
\begin{align}
    \sigma^2_{\text{GMM}} &= \frac{1}{K}\sum_{k=1}^K \frac{\textrm{tr}(\hat{\Sigma}_k) - \lambda_1(\hat{\Sigma}_k)}{d-1}, \label{eq:sigma_gmm}\\
     \x_{\text{GMM},k} &= \sqrt{\left(\lambda_1(\hat{\Sigma}_k)-\sigma^2_{\text{GMM}}\right)} \: \vb_1(\hat{\Sigma}_k) &\forall k \in [K]. \label{eq:xk_gmm}
\end{align}
Note that the requirement in \eqref{eq:larg_spectral_gap} can be thought of as a large enough spectral gap for the leading eigenvalues of $\Sigma_k$, 
and it has been checked to hold in all the following experiments of this section.

\begin{figure*}[t]
\centering
\subfloat[Low noise $\sigma^2=0.01$]{\includegraphics[width=0.9\textwidth]{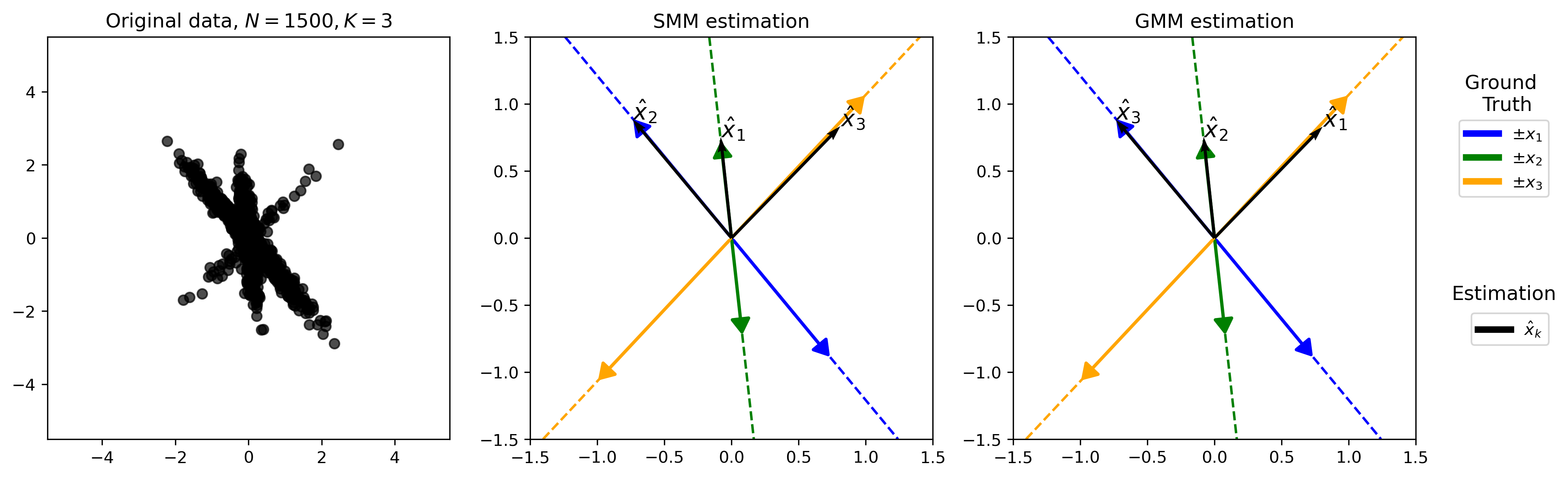}\label{fig:gmm_comparison_a}}
\hspace{10cm}
\newline\newline
\subfloat[High noise $\sigma^2=0.5$]
{\includegraphics[width=0.9\textwidth]{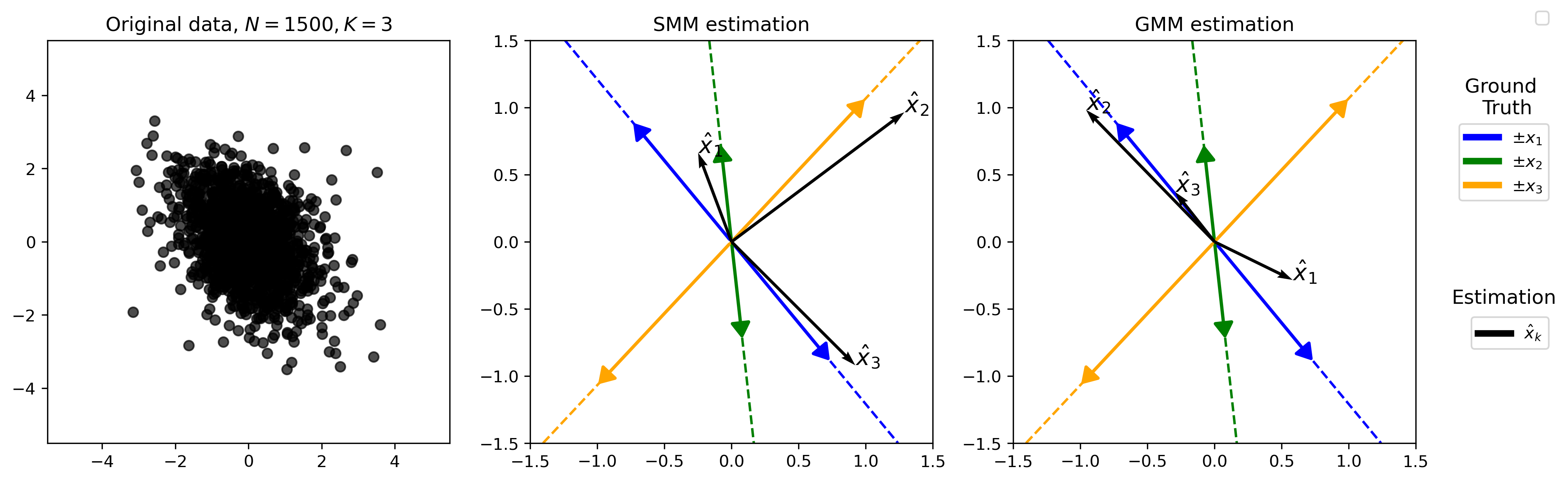}\label{fig:gmm_comparison_b}}
\caption{Comparison between recovery by \ac{GMM} versus \ac{SMM}. The comparison is made at both low noise ($\sigma^2 = 0.01$) \eqref{fig:gmm_comparison_a} and high noise ($\sigma^2 = 0.5$) \eqref{fig:gmm_comparison_b}. The fixed parameters are: $\x_1 \approx [0.75,-0.91], \x_2 \approx [0.08, -0.75], \x_3 \approx [-1.01,-1.08]$, $\pi_1 \approx 0.58, \pi_2 \approx 0.37, \pi_3 \approx 0.05$.}
\label{fig:gmm_comparison}
\end{figure*}

In what follows, we compare the performance of \ac{SMM} and \ac{GMM}-based recovery on a synthetic dataset.\footnote{We remark in passing that this setup models, \textit{e.g.}, a random access SIMO channel, where various sensors with one antenna sporadically transmit a symbol to a base station with many antennas at random times. The algorithm blindly estimates both the channel vectors $\x_k$ and the symbols $\alpha$.}
Specifically, we choose $K=3$, and sample fixed (starred) parameters:
\begin{empheq}[left=\empheqlbrace]{equation}
    \begin{aligned}
     \vtheta^\star &= \{\x_1^\star , \ldots , \x_K^\star, \pi_1^\star, \ldots , \pi_K^\star , {\sigma^2}^{\star} \}, \\
     \pi_k^\star &\in [0,1], \: \sum_{k=1}^K \pi_k^\star =1 \\
    {\sigma^2}^{\star} &\in \mathbb{R}_+.
    \end{aligned}
    \label{eq:synthetic}
\end{empheq}
Given these parameters, we generate $N=1500$ samples, $\y_1,\ldots, \y_N$, according to the model in \eqref{eq:model}, and we do this for two different noise levels: $\sigma^2 = 0.01$ and $\sigma^2 = 0.5$.
Our goal is to assess how the noise level affects the accuracy of recovering the true signal vectors $\x_1$, $\x_2$, and $\x_3$ using \ac{SMM} and \ac{GMM}.
We emphasize once more that, due to symmetry, the recovery of $\x_k$ is just as likely as the recovery of $-\x_k$.
The results are shown in Figure~\ref{fig:gmm_comparison}.
While in the low noise case (Figure~\ref{fig:gmm_comparison_a}), the performance of \ac{SMM} and \ac{GMM} appear to be similar, we see a clear difference for the high noise case (Figure~\ref{fig:gmm_comparison_b}).
 In this regime, the accuracy of both methods decreases, but \ac{GMM} only finds one vector, $\x_1$, and produces a
third estimate, which is a mixture of two ground truth signals. On the other hand, \ac{SMM} delivers three clearly separate directions without the 'collapse' of estimates we see in the \ac{GMM} case.

To quantify \ac{SMM}'s performance versus \ac{GMM}'s, we conducted two experiments. One compares the methods' noise estimation performance, and the other compares the distance of the estimated signals, $\hat{\mathcal{X}} = \{\hat{\x}_k\}_{k=1}^K$, to the true ones, $\mathcal{X} = \{\x^\star_k\}_{k=1}^K$.
\textcolor{manualcolor}{
The supplementary material, section~\ref{sec:convergence_speed}, furthermore contains an empirical experiment that suggests that the convergence speed of both algorithms is similar.
}

\subsection{Noise estimation}
\label{sec:noise_estimation}
\begin{figure}[t]
\centering
\includegraphics[width=2.9in]{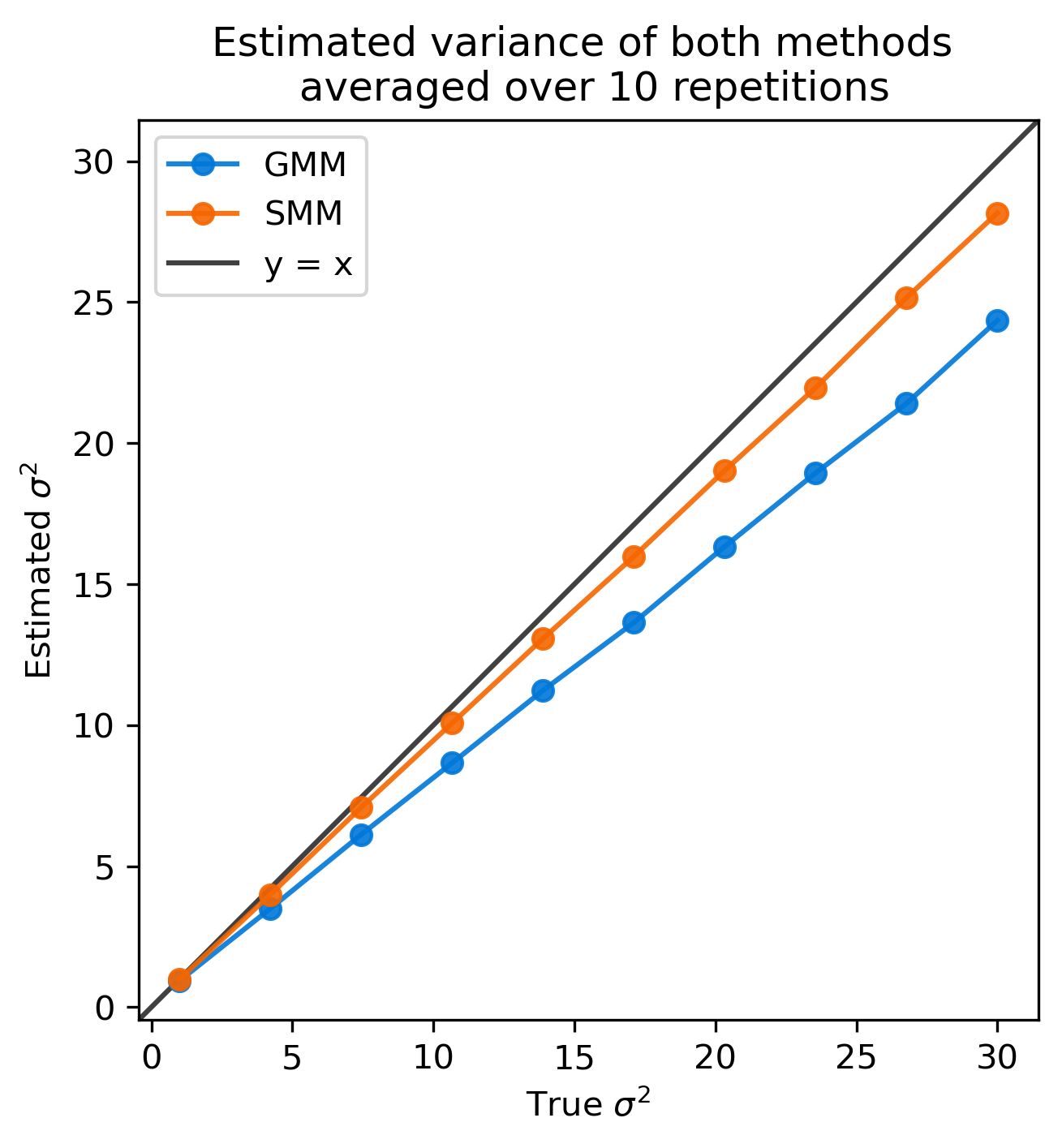}
\caption{Comparison of \ac{SMM}’s and \ac{GMM}’s noise variance estimation versus the ground truth value, at $10$ different noise levels equally spaced in $[1,30]$. At each noise level, 10 replicate datasets with different underlying signals are generated using $N=1500$ samples and the parameters $d=5$ and $K=3$.}
\label{fig:bias}
\end{figure}

\noindent
We conducted signal recovery using both methods on synthetic datasets with a known ground truth noise variance $\sigma^2$, repeating the process across $10$ different noise levels in the range $[1,30]$.
Figure~\ref{fig:bias} shows the results of \ac{SMM}'s and \ac{GMM}'s noise variance estimation compared to the ground truth values.
While both methods tend to underestimate the noise variance, the bias is significantly larger for \ac{GMM} than for \ac{SMM}.
Additionally, the norms of the estimated $\hat{\x}_k$ decrease as $\sigma^2$ increases (see \eqref{eq:lambda},\eqref{eq:xk_gmm}), leading to overestimated norms for both methods. This overestimation is notably more pronounced in the \ac{GMM} case, which further exacerbates its bias.

\subsection{Signal estimation}
\begin{figure}[t]
\centering
\includegraphics[width=3in]{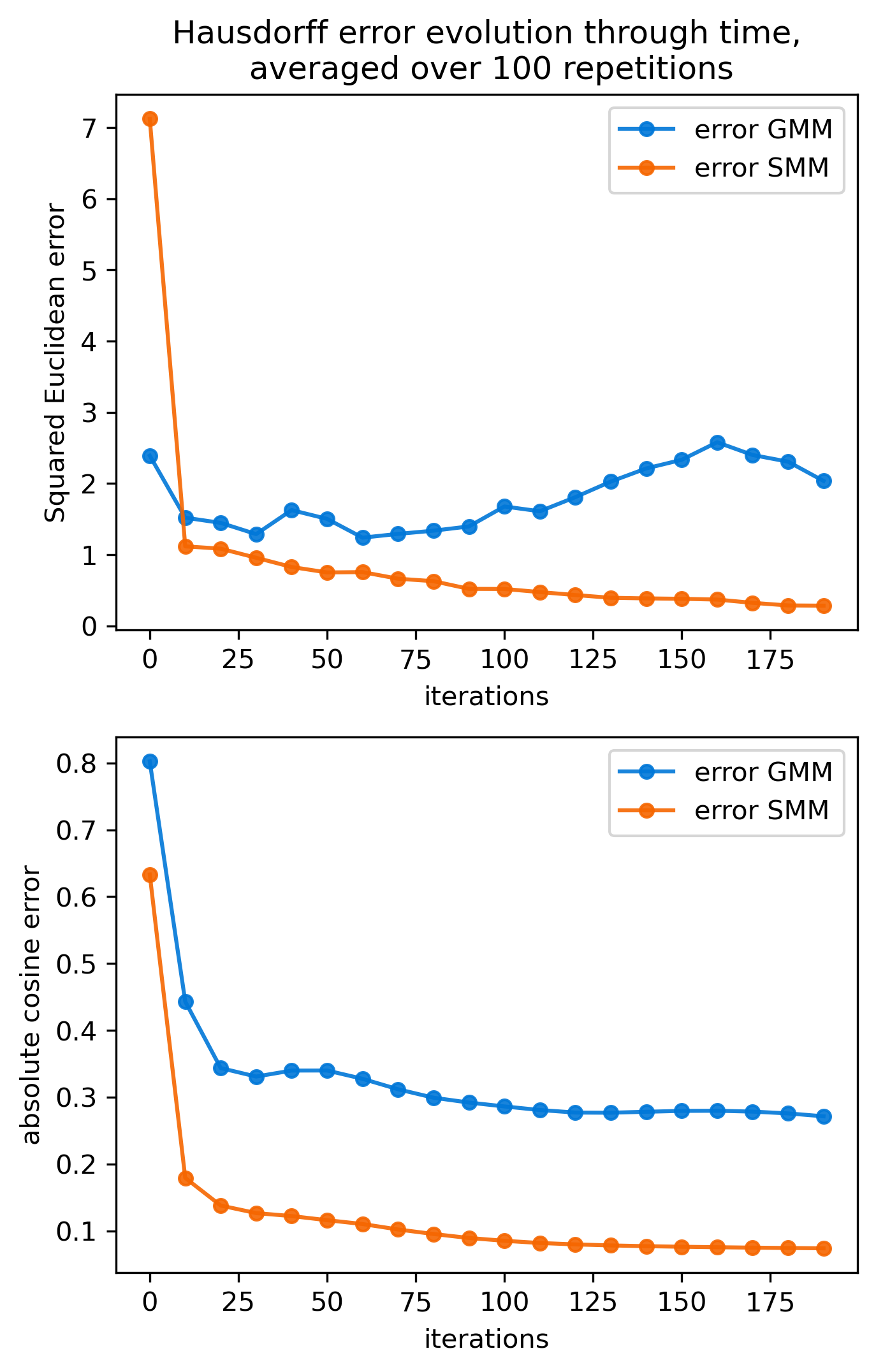}
\caption{
Comparison of \ac{SMM}’s and \ac{GMM}’s estimated signals versus the ground truth signals, using a synthetic dataset with $N=1500$, $K=3$, $d=5$, $\sigma^2= 1.5$, $\pi_1 \approx 0.62$, $\pi_2 \approx 0.22$, and $\pi_3 \approx 0.16$. We repeat the experiment for 100 different initializations of both methods and report the average error at every iteration.
The Hausdorff distance between the estimated vectors $\{\hat{\x}_k\}_{k=1}^K$ and true vectors $\{\x_k\}_{k=1}^K$, for both the squared Euclidean distance and the absolute cosine distance, is consistently smaller for \ac{SMM} compared to \ac{GMM}, suggesting better signal recovery in the \ac{SMM} case.}
\label{fig:hausdorff_distance}
\end{figure}

\noindent
To assess the fidelity of recovered signals, we performed estimation on a synthetic dataset with parameters $N=1500$, $K=3$, $d = 5$, and $\sigma^2 = 1.5$.
The ground truth signals $\x$ were compared to their estimated counterparts $\hat{\x}$ using the squared Euclidean and absolute cosine error distance metrics: 
\begin{align*}
    d_{\textrm{sqe}}(\x, \hat{\x}) &:= \|\x-\hat{\x}\|_2^2, \\    
    d_{\textrm{abs\_cos}}(\x, \hat{\x}) &:= 1-\frac{|\x^\mathrm{T} \hat{\x}|}{\|\x\|_2\|\hat{\x}\|_2}.   
\end{align*}    
The squared Euclidean distance, $d_{\text{sqe}}(\x,\hat{\x})$, depends on the norms of $\x, \hat{\x}$, while $d_{\text{abs\_cos}}(\x,\hat{\x})$ does not.
Since Section~\ref{sec:noise_estimation} highlighted a bias in the estimated norms, $d_{\text{abs\_cos}}(\x,\hat{\x})$ was included for its invariance to such bias.
To quantify the overall discrepancy between the set of estimated signals $\hat{\mathcal{X}}$ and the set of true signals $\mathcal{X}$, we used the Hausdorff distance.
For a distance metric $d(\cdot)$, the Hausdorff distance between sets $\mathcal{X}$ and $\hat{\mathcal{X}}$ is defined as: 
\begin{align*}
    &d_H\left(\mathcal{X},\hat{\mathcal{X}}\right) := \max \left(\sup_{\x \in \mathcal{X}} d(\x,\hat{\mathcal{X}}) , \sup_{\hat{\x} \in \hat{\mathcal{X}}} d(\mathcal{X},\hat{\x})\right),\textrm{ where}\\
    &d(\x,\hat{\mathcal{X})} = \inf_{\hat{\x}\in \hat{\mathcal{X}}} d(\x,\hat{\x})\quad\textrm{ and }\quad
    d(\mathcal{X},\hat{\x}) = \inf_{\x \in \mathcal{X}} d(\x,\hat{\x}).
\end{align*}
The results in Figure~\ref{fig:hausdorff_distance} show that the Hausdorff distance between estimated signals $\hat{\mathcal{X}}$ and true signals $\mathcal{X}$ is substantially smaller for \ac{SMM} compared to \ac{GMM}, regardless of whether $d_{\text{sqe}}(\x,\hat{\x})$ or $d_{\text{abs\_cos}}(\x,\hat{\x})$ is used.
These findings suggest that \ac{SMM} achieves more accurate signal recovery than \ac{GMM}.

\section{Applications}
\label{sec:applications}
\noindent
We demonstrate \ac{EM}-based fitting of a \ac{SMM} in two practical applications.
Given \eqref{eq:model}, the \ac{SMM} is particularly well suited for measurements in which an observation is underlain by a relative signal, this signal is randomly scaled to an absolute signal by a process not under the control of the measurer, and the observation is also perturbed by external noise.
Although this might seem rather particular at first, it is a remarkably common signal structure carried by quite different measurement types in different domains.
For example, a biological cell has a particular relative abundance profile of the molecular species it contains.
When measured by mass spectrometry, a mass spectrum can report that abundance profile, but measured absolute intensities will be scaled by the chemical matrix, \textit{i.e.}, the overall chemical environment present at that measurement location.
Similarly, a sensor aimed at an object to record its color can report an electromagnetic spectrum profile, but that relative profile can be randomly scaled, \textit{e.g.}, by atmospheric circumstances, before it reaches the sensor.

Here, we test \ac{SMM}-fitting on an \acl{IMS} and a \acl{HSI} dataset.
Specifically, we use \ac{SMM} to recover underlying molecular signatures from
noisy \ac{IMS} measurements of a rat brain tissue section, and to estimate underlying color spectra from the \ac{HSI} measurement of a scene picturing the Statue of Liberty.
In both applications, feature-wise min-max normalization is employed to ensure all features have equal weight.

\ac{EM} is typically focused on estimating subpopulations of a dataset.
However, the E-Step also computes responsibility variables $\rho_{k}^{[i]}$ that tie an observation $i$ to a subpopulation $k$, and that can be used to implicitly cluster the data.
As clustering pixel measurements is equivalent to segmenting an image, we can use the implicit segmentation results to assess the quality of the estimated subpopulations (or spikes) by means of their corresponding image segment.
In these examples, we choose to cluster according to maximum probability.
Namely, an observation $i\in[N]$ is associated to cluster $c_i$, where
\begin{align*}
    c_i := \argmax_{c\in[K]} \rho_{c}^{[i]}.
\end{align*}
Furthermore, we compare \ac{SMM}'s clustering results to ones given by traditional methods such as \ac{GMM} and \acf{KMC}. 

Note that at every step of Algorithm~\ref{alg:em}, we need to compute matrices $\A_k = \sum_{i=1}^N \rho_e^{[i]} \y_i \y_i^\mathrm{T}$, $\forall k \in [K]$. Without parallelization, this takes $O(N)$ and can significantly slow down the full algorithm. 
The following lemma shows that the process can be sped up using an approximate matrix $\tilde{\A}_k$, preserving the leading eigenvalue up to a controlled precision.
\begin{lemma}
    For every $\delta > 0$, let $\tau := \delta /\|\Y\|_\mathrm{F}^2$. The matrix
    \begin{align*}
        \Tilde{\A}_k = \sum_{\substack{i=1 \\ \rho_k^{[i]} \geq \tau}}^N \rho_k^{[i]}\y_i\y_i^\mathrm{T}\quad\textit{ satisfies }\quad |\lambda_1(\A_k) - \lambda_1(\Tilde{\A}_k)| \leq \delta.
    \end{align*}     
    \label{lem:approximated_Ak}
\end{lemma}
\noindent
The proof of this lemma uses Weyl's inequality and is provided in Section~\ref{sec:proof_results} of the Supplementary Material.
Note that for $\tau\geq 0$ large enough, computing $\Tilde{\A}_k$ is much faster since we only need to sum over the indices $i$ satisfying $\rho_e^{[i]} \geq \tau$.

\begin{table}[b]
\centering
\begin{tabular}{ |r||c|c|}
 \hline
 &Rat brain \ac{IMS} & Salient \ac{HSI} \\
 \hline
 Image size (pixels) & $1404 \times 408$ &  $1024 \times 728$ \\
 \hline
 Nr. of spectral bands & $843$ & $81$  \\
 \hline
 Spectral range & $[403.245,1573.905]$ Da & $[380,780$] nm \\
 \hline
\end{tabular}
\newline
\caption{Dataset specifications}
\label{tab:dataset_spec}
\end{table}

\subsection{Results on \acl{IMS}}
\begin{figure*}[t]
\centering
\includegraphics[width=0.85\textwidth]{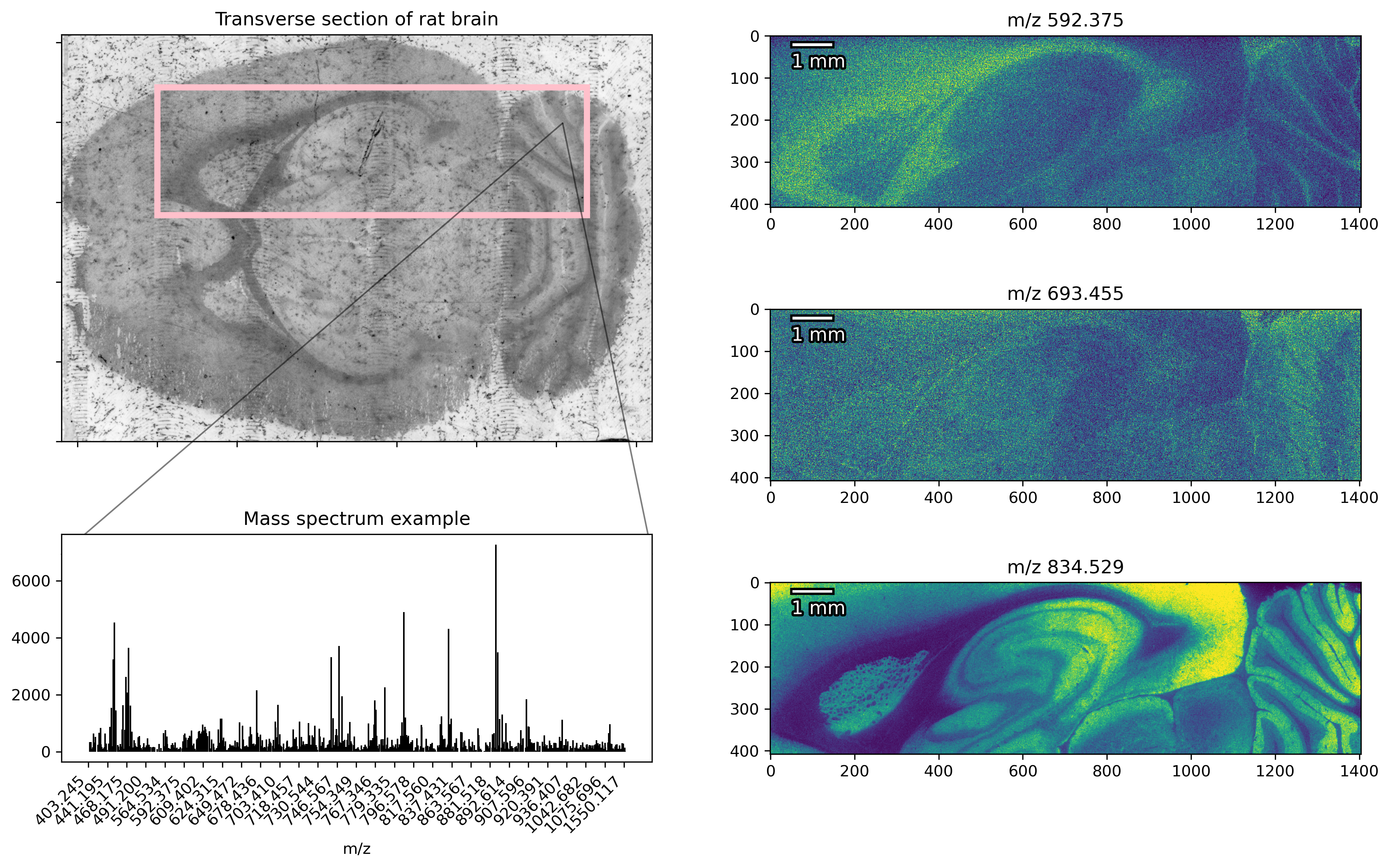}
\caption{Rat brain \ac{IMS} dataset.
(top-left) A transverse section of rat brain tissue was analyzed using \ac{IMS} (analysis area outlined in pink).
(right) Three of the 843 ion images acquired, showing the spatial distributions of ion species corresponding to \ac{mz} $592.375$, $693.455$, $834.529$.
(bottom-left) An example mass spectrum (peak-picked) acquired at a single pixel, reporting $843$ \ac{mz} features.}
\label{fig:ims_data}
\end{figure*}

\noindent
\Acf{IMS} \cite{caprioli1997molecular,colley2024high} is a molecular imaging technique that combines spatial mapping with (mass) spectral analysis.
It offers detailed chemical maps of samples such as organic tissues or biofilms, measuring the distributions of hundreds of molecular species throughout a defined measurement region.
An \ac{IMS} measurement can typically be considered as a 3-mode tensor (two spatial modes and one spectral mode), where every entry reports the ion intensity at a specific spatial location and a particular \ac{mz} value.
It provides a (gray scale) ion image for each recorded \ac{mz} value, visualizing where the compound corresponding to that \ac{mz} value is located in the sample.
Each pixel implicitly records a full mass spectrum or \ac{mz} profile, revealing which molecular compounds are present at that particular location. 
Since prior labeling of compounds is not required and a single measurement simultaneously reports hundreds of molecular species, \ac{IMS} has become an important imaging modality for the molecular exploration of the content of biological tissues and for elucidation of disease-related mechanisms.

In this study, a transverse section of a rat brain was measured using \ac{QTOF} \ac{IMS}.
After preprocessing (Supplementary Material, Section~\ref{app:rat_brain_acquisition}), we obtain an \ac{IMS} dataset with the specifications listed in Table~\ref{tab:dataset_spec}.
Figure~\ref{fig:ims_data} shows three of the 843 ion images in this dataset, which depict the distributions of three different molecular species, alongside an example mass spectrum acquired at a particular location, reporting localized abundances for 843 distinct ions there.

\begin{figure}[t]
\centering
\begin{minipage}[b]{0.5\textwidth}
    \centering
    \includegraphics[width=\textwidth]{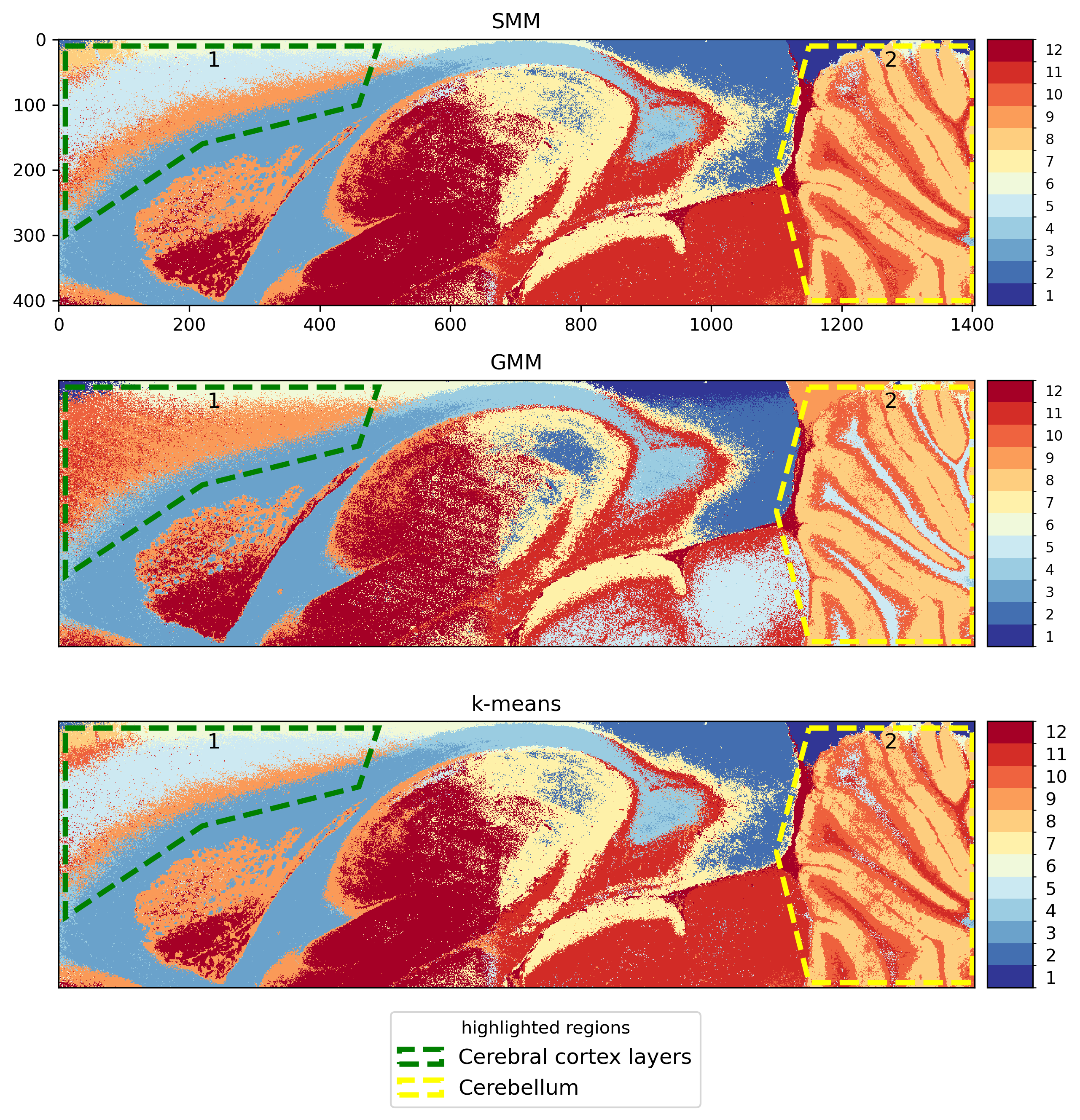}
\end{minipage}
\caption{Clustering results on the rat brain \ac{IMS} dataset with $k=12$, for \ac{SMM}, \ac{GMM}, and \ac{KMC}.
These results demonstrate SMM’s ability to retrieve signals in low-SNR environments. 
(region 1 – cerebral cortex) SMM discerns a biological subdivision of the cortex layers that is known to exist, but that is missed by GMM.
(region 2 – cerebellum) SMM exhibits less susceptibility to noise than \ac{KMC}.}
\label{fig:ims_results}
\end{figure}

Figure~\ref{fig:ims_results} compares the clustering outputs obtained by \ac{SMM}-fitting, \ac{GMM}-fitting, and \acl{KMC}, all for $k=12$ (results for other $k$ in the Supplementary Material).
Given that many of the ion images are relatively noisy (see \ac{mz} 592.375 and 693.455 ion images in Fig.~\ref{fig:ims_data}), the results demonstrate \ac{SMM}'s ability to retrieve signals in low-\ac{SNR} environments.
The potential of \ac{SMM}-based signal recovery is particularly illustrated in the highlighted areas of Fig.~\ref{fig:ims_results}, where \ac{SMM} is able to recover biological patterns that are missed by other methods.
Specifically, in region~$1$ (the cerebral cortex), \ac{SMM} provides a subdivision of the cortex layers (\textit{e.g.}, molecular, granular, pyramidal, and multiform) that are known to align parallel to the surface of the brain, but that is missed by \ac{GMM}.
In region~$2$ (the cerebellum), \ac{SMM} delivers sharper delineation of key anatomical structures (\textit{e.g.}, white matter, molecular layer, and granule cell layer) and exhibits less susceptibility to noise than \ac{KMC}.
While we do not claim that \ac{SMM} is superior in all cases, the difference in performance in this example is substantial enough to consider this method.
Furthermore, in Fig.~\ref{fig:spectra_comparision_rat} in the Supplementary Material, we compare the estimated subpopulation spectra $\{\hat{\x}_k\}_{k=1}^K$ for \ac{SMM} and \ac{GMM}.
The signals estimated by \ac{SMM} seem closer to real mass spectra than \ac{GMM}-estimated signals.
For example, \ac{IMS} data is inherently non-negative.
Without explicitly imposing non-negativity, \ac{SMM}-recovered signals are largely non-negative and more similar to real mass spectra than the \ac{GMM}-recovered signals, which exhibit substantial amounts of negative values.
\subsection{Results on \acl{HSI}}
\begin{figure}[t]
\centering
\begin{minipage}[b]{0.23\textwidth}
    \centering
    \includegraphics[width=\textwidth]{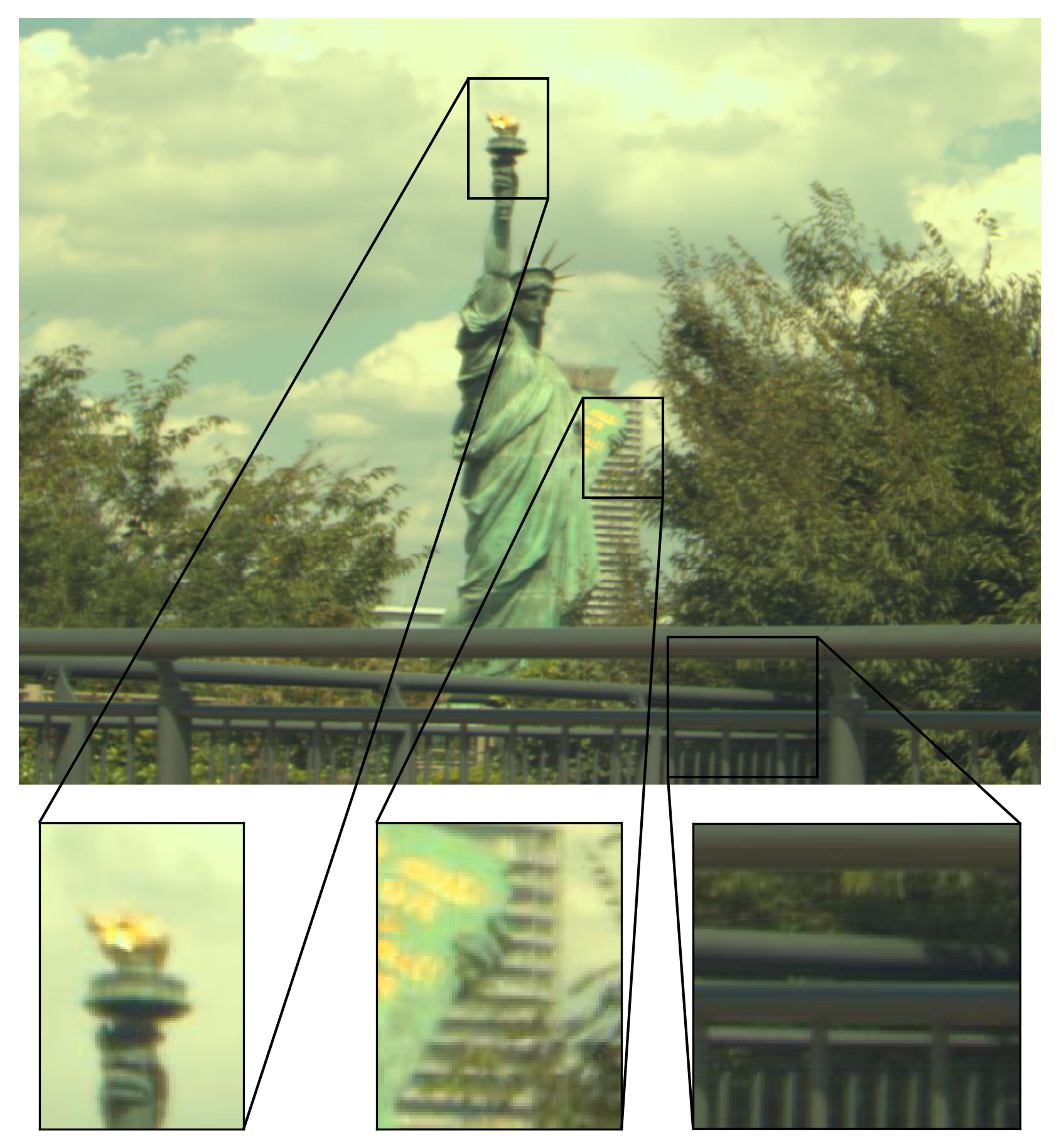}
\end{minipage}%
\caption{RGB reference image for the \ac{HSI} dataset \cite{salient_data}.}
\label{fig:HSI_RGB}
\end{figure}

\noindent
\Acf{HSI}, like \ac{IMS}, provides both spatial and spectral information, albeit of a different scale and nature with \ac{HSI} reporting electromagnetic wavelengths.
This study uses the salient object dataset introduced in \cite{salient_data}, which consists of hyperspectral images of well-known objects captured under various spectral conditions.
The dataset was designed to evaluate the performance and robustness of models in detecting and segmenting salient objects.
However, we leverage it here because the imaged objects are familiar to most readers (\textit{e.g.}, the Statue of Liberty, Fig.~\ref{fig:HSI_RGB}), enabling easy interpretation of what constitutes an improvement in segmentation.
Details of the \ac{HSI} dataset can be found in Table~\ref{tab:dataset_spec}.

\begin{figure*}[t]
    \centering
    \includegraphics[width=0.9\textwidth]{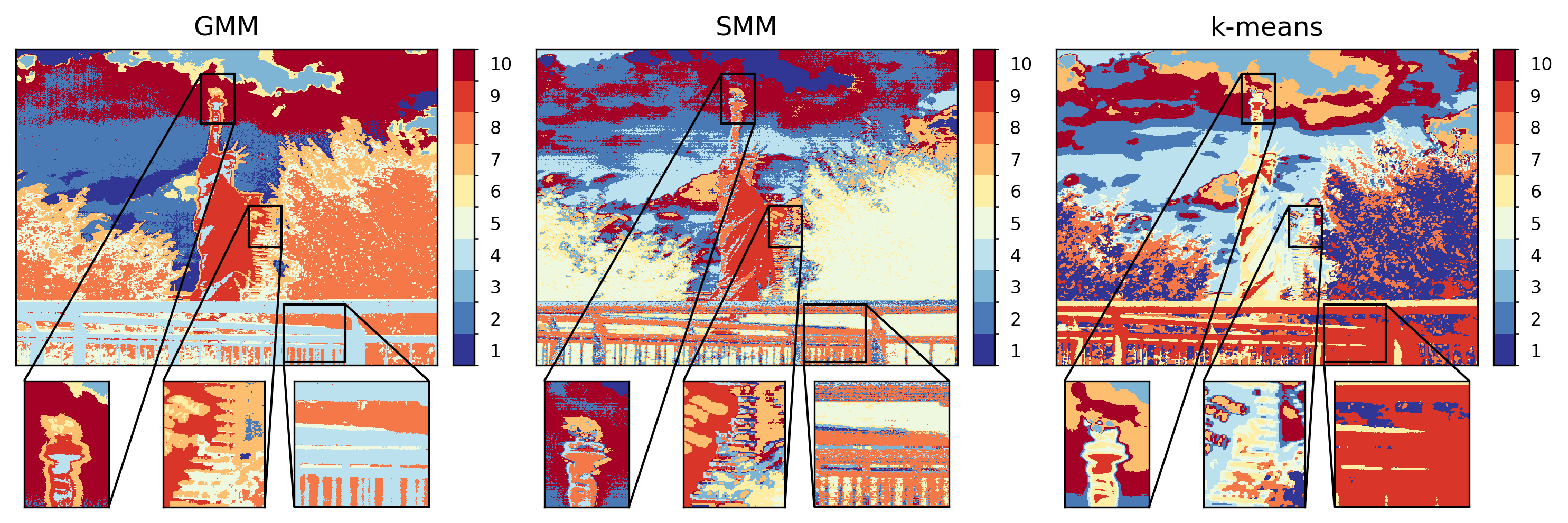}
    \caption{Clustering results on the \ac{HSI} dataset \cite{salient_data} with $k=10$, for \ac{GMM} (left), \ac{SMM} (middle), and \ac{KMC} (right).
These results demonstrate SMM’s ability to discern different underlying signals. 
(torch) \Ac{SMM} separates the flame from the torch handle and, unlike \ac{GMM} and \ac{KMC}, avoids the `halo'-type signal surrounding the torch and flame.
(tablet) The tablet is recovered by \ac{SMM} as a signal distinct from the building behind it, while \ac{GMM} and \ac{KMC} both exhibit spill-over between the tablet and building signals.
(railing) \Ac{KMC} has difficulty discerning the railing from other dark areas such as the foliage, while \ac{SMM} and \ac{GMM} successfully separate the foliage and railing signals.
\Ac{SMM} furthermore surpasses \ac{GMM} in delivering more railing sub-signals, separating out areas with differing amounts of sunlight and shadow (see RGB image in Fig.~\ref{fig:HSI_RGB} for reference).}
\label{fig:salient_model_comparison}
\end{figure*}

As in the previous section, Figure~\ref{fig:salient_model_comparison} compares \ac{SMM}, \ac{GMM}, and \ac{KMC} results for $k=10$ (results for other $k$ in the Supplementary Material).
Figure~\ref{fig:HSI_RGB} provides an RGB image for reference, with three regions of interest highlighted: the torch and flame, the tablet in the hand, and the metal railing in the foreground. 
All three reveal noticeable differences between the methods, described in the caption of Fig.~\ref{fig:salient_model_comparison}.
In these examples, \ac{SMM} exhibits a remarkable ability to discern details that \ac{GMM} and \ac{KMC} might not differentiate and instead view as a single signal.
The estimated signal spectra are also provided in Fig.~\ref{fig:min_max_salient_min_max} of the Supplementary Material.
While \ac{SMM} fitting was not initially intended as a clustering method, its ability to differentiate signal subpopulations that are not discerned by approaches such as \ac{GMM} and \ac{KMC} is a testament to its ability for robust signal recovery in a noisy environment.


\section{Conclusion}
\noindent
We introduced the \ac{SMM} and a corresponding \ac{EM} algorithm as a new method for signal recovery.
Although it is more restrictive, for data types where the \ac{SMM} applies, \ac{SMM} offers substantially better recovery than \ac{GMM} in low-\ac{SNR} regimes. \textcolor{manualcolor}{While this paper makes a comparison to the standard \ac{GMM}, one potential future direction could be to compare to more robust versions of \ac{GMM} \cite{tadjudin2002robust}}.
Applications to real-world \ac{IMS} and \ac{HSI} data demonstrate more accurate signal recovery than with traditional methods and provide a powerful spatial clustering method, lifting data features from the noise that might otherwise go unrecognized.
\textcolor{manualcolor}{
Another interesting direction for future work could be to generalize our results to settings where the noise and the random scaling distributions have heavier tails, such as the Student’s $t$ or generalized hyperbolic distributions, which may enhance robustness to outliers. Such a model could construct an even more robust finite mixture model \cite{murray2017hidden} \cite{lee2016finie} \cite{peel2000robust} \cite{browne2015mixture}}.

\section{Acknowledgment}
\noindent
Research was supported by the National Institutes of Health (NIH)’s NIDDK (U54DK134302 and U01DK133766), NEI (U54EY032442), NIAID (R01AI138581 and R01AI145992), NIA (R01AG078803), NCI (U01CA294527), and by the Chan Zuckerberg Initiative DAF (2021-240339 and 2022-309518). The content is solely the responsibility of the authors and does not necessarily represent the official views of the NIH.

\begin{appendices}
\section{Proofs of section~\ref{sec:expected_maximization}}
\label{sec:two}
\begin{lemma}
\label{lem:sigma_positive}
    For fixed values of $\rho_e^{[i]}$ and $S\subseteq [K]$, the estimated variance of the noise
    \begin{align*}
        \sigma^2 = \frac{\|\Y\|_\mathrm{F}^2 -\sum_{k \in S} \lambda_k}{dN - \sum_{k \in S}\gamma_k } \quad\quad \textrm{is positive.}
    \end{align*}
\end{lemma}
\begin{proof}
    For $d\geq 2$, the denominator is positive since $\sum_{k\in S}\gamma_k \leq N$.
    For $\sigma^2$ to be proven positive, we only need to prove that the numerator is positive.
    Since $\A_k \succeq 0$, we get:
    {\allowdisplaybreaks
    \begin{align*}
        \lambda_k &\leq \textrm{tr}(\A_k)  &\forall k \in \{1,\ldots , K\}\\
        \sum_{k \in S} \lambda_k &\leq \sum_{k\in S}\textrm{tr}(\A_k) &\textrm{Summing over }k \in S \\
        &\leq \sum_{k=1}^K \textrm{tr}(\A_k) &S \subseteq [K], \textrm{tr}(\A_k)\geq 0\\
        &= \textrm{tr}\left(\Y \sum_{k=1}^{K} \textrm{diag}(\rho_k) \: \Y^\mathrm{T} \right) &\textrm{Using } \A_k \textrm{ from } \eqref{eq:Ae} \\
        &= \textrm{tr}\left(\Y \I \Y^\mathrm{T} \right) &\forall i\in[N], \sum_{k=1}^K \rho_k^{[i]} = 1 
        \\
        &= \|\Y\|_\mathrm{F}^2
    \end{align*}
    }
\end{proof}

\subsection{Expression of $\mathcal{Q}(\vtheta;\vtheta^{[t]})$ at a critical point}
\label{sec:app_Q_saddle_point}
\noindent
The following intermediate steps allow simplification of the expression of $\mathcal{Q}(\vtheta;\vtheta^{[t]})$ at a critical point $\vtheta$.
In Section~\ref{sec:expected_maximization}, equations for a critical point were obtained $\forall k \in \{1,\ldots K\}$:
{\allowdisplaybreaks
\begin{align}
    \pi_k &= \frac{\gamma_k}{N} \label{eq:pi_app}  \\ 
    &\left\{
    \begin{array}{cc}
    \textrm{either }\x_k = \vnull & \\ 
    \textrm{or }\A_k = \lambda_k \x_k, &\|\x_k\|^2 = \frac{\lambda_k}{\gamma_k} - \sigma^2 
    \end{array}\right. \label{eq:eig_val_app}\\
    \sigma^2 &= \frac{ \|\Y\|_\mathrm{F}^2 -\sum_{k : \x_k \neq \vnull} \lambda_k}{d N - \sum_{k: \x_k \neq \vnull} \gamma_k} \label{eq:sigma_app}
\end{align}
}
Using \eqref{eq:eig_val_app} and adding a zero, for $\x_k \neq \vnull$, we get
\begin{align}
    \frac{\x_k^\mathrm{T} \A_k \x_k}{\|\x_k\|^2 + \sigma^2} 
    &= \lambda_k \frac{\|\x_k\|^2 + \sigma^2 - \sigma^2}{\|\x_k\|^2 + \sigma^2} \nonumber \\
    &= \lambda_k - \sigma^2 \frac{\lambda_k}{\|\x_k\|^2 + \sigma^2}
    &= \lambda_k - \sigma^2 \gamma_k .
    \label{eq:combine_crit_eq}
\end{align}
For $\vtheta$ satisfying \eqref{eq:pi_app}, $\mathcal{Q}(\vtheta;\vtheta^{[t]})$ in \eqref{eq:Q_function} can be written as
\begin{align*}
    \mathcal{Q}(\vtheta; \vtheta^{[t]}) &= C -\frac{\|\Y\|_\mathrm{F}^2}{2\sigma^2} + \frac{1}{2\sigma^2}\sum_{k=1}^K\frac{\x_k^\mathrm{T} \A_k \x_k}{\|\x_k\|^2 + \sigma^2} \\
    & -\frac{1}{2} \sum_{k=1}^K \gamma_k \ln(\|\x_k\|^2+\sigma^2) -\frac{d-1}{2}N \ln(\sigma^2),
\end{align*}
where $C = \sum_{k=1}^K \gamma_k \ln \frac{\gamma_k}{N} -\frac{dN}{2} \ln(2\pi)$, a constant independent of $\vtheta$.

\noindent
Using \eqref{eq:combine_crit_eq} and \eqref{eq:eig_val_app}, we can simplify further:
{\allowdisplaybreaks
\begin{align*}
    &\mathcal{Q}(\vtheta; \vtheta^{[t]}) \\
    &= C - \frac{\|\Y\|_\mathrm{F}^2}{2\sigma^2} + \frac{1}{2\sigma^2}\sum_{k:\x_k \neq \vnull} [\lambda_k - \sigma^2 \gamma_k]  \\
    &\quad -\frac{1}{2}\sum_{k : \x_k\neq \vnull} \gamma_k \ln \frac{\lambda_k}{\gamma_k} 
     -\frac{1}{2} \left[(d-1)N + \sum_{k : \x_k =\vnull} \gamma_k \right] \ln \sigma^2 \\
    &=C -\frac{1}{2\sigma^2} \left[\|\Y\|_\mathrm{F}^2 - \sum_{k : \x_k \neq \vnull} \lambda_k\right]
     -\frac{1}{2} \sum_{k : \x_k \neq \vnull} \gamma_k \\ 
    &\quad -\frac{1}{2} \sum_{k: \x_k \neq \vnull} \gamma_k \ln \frac{\lambda_k}{\gamma_k} -\frac{1}{2} \left[(d-1)N + \sum_{k : \x_k =\vnull} \gamma_k \right] \ln \sigma^2. \\
\end{align*}
}
Using \eqref{eq:sigma_app} and $\sum_{k=1}^K \gamma_k = N$ in the last term, yields
{\allowdisplaybreaks
\begin{align*}
    &\mathcal{Q}(\vtheta; \vtheta^{[t]}) \\    
    &= C -\frac{1}{2} \left[dN - \cancel{\sum_{k:\x_k \neq \vnull} \gamma_k}\right]-\frac{1}{2} \cancel{\sum_{k : \x_k \neq \vnull} \gamma_k} \\
    &\quad  -\frac{1}{2} \sum_{k: \x_k \neq \vnull} \gamma_k \ln \frac{\lambda_k}{\gamma_k} -\frac{1}{2} \left[dN - \sum_{k : \x_k \neq \vnull} \: \gamma_k \right] \ln \sigma^2  .
\end{align*}
}
Finally, denoting $C_2 = C -\frac{dN}{2}$, we get 
\begin{align*}
    \mathcal{Q}(\vtheta; \vtheta^{[t]}) 
    &= C_2 -\frac{1}{2} \left[dN - \sum_{k : \x_k \neq \vnull} \gamma_k \right] \ln \sigma^2 \\
    &\quad -\frac{1}{2} \sum_{k: \x_k \neq \vnull} \gamma_k \ln \frac{\lambda_k}{\gamma_k}. 
\end{align*}

\section{Greedy optimizer for critical point}
\label{sec:appendix_greedy_opt}
\noindent
We show that a greedy optimizer solves the optimization problem
\begin{align}
    \begin{array}{lll}
         S^\star = & \text{argmin} & g(S) \\ 
         & S \in \mathcal{V}
    \end{array}
    \label{eq:optimization_problem}
\end{align}
\begin{align*}
    \textrm{with }\quad \sigma^2(S) &= \frac{\|\Y\|^2_\mathrm{F} - \sum_{k\in S} \lambda_k}{dN - \sum_{k \in S}\gamma_k}, \\
    g(S) &= \left[dN - \sum_{k\in S} \gamma_k\right] \ln \left(\frac{\|\Y\|^2_\mathrm{F} - \sum_{k\in S} \lambda_k}{dN - \sum_{k \in S} \gamma_k}\right) \\
    &\quad + \sum_{k\in S} \gamma_k\ln \left(\frac{\lambda_k}{\gamma_k}\right),\\
    \mathcal{V} &= \left\{ S\subseteq [K] ;\: \forall k \in S : \sigma^2(S) \leq \lambda_k/ \gamma_k \right\}.
\end{align*}
Lemma~\ref{lem:decrease_set} guarantees that
removing elements from a valid set only increases the objective function of \eqref{eq:optimization_problem}.
\begin{lemma}
\label{lem:decrease_set}
    For $S\subseteq[K]$ and $j \in S$ where $\sigma^2(S) \leq \lambda_j/\gamma_j$,
    \begin{align*}
        \textrm{we have: }\quad\quad g(S) \leq g(S\setminus\{j\}).
    \end{align*}
\end{lemma}
\begin{proof}
    From the definitions of $\sigma^2(S)$ and $\sigma^2(S\setminus\{j\})$, we find    
    \begin{align}
        \sigma^2(S) = \frac{\sigma^2(S\setminus\{j\})[dN - \sum_{k\in S} \gamma_k + \gamma_j] - \lambda_j}{dN - \sum_{k\in S}\gamma_k }
        \label{eq:sigma_new}.
    \end{align}
    \noindent
    We will show that $g(S\setminus\{j\}) - g(S)\geq 0$.
    {\allowdisplaybreaks
    \begin{align*}
        &g(S\setminus\{j\}) -g(S) \\
        &= \left[dN - \sum_{k\in S} \gamma_k +\gamma_j \right] \ln \sigma^2(S\setminus\{j\}) + \sum_{k\in S\setminus\{j\}}\gamma_k \ln \frac{\lambda_k}{\gamma_k}\\
        &\:\:\:\: -\left[dN -\sum_{k \in S}\gamma_k \right] \ln\sigma^2(S) -\sum_{k \in S} \gamma_k \ln \frac{\lambda_k}{\gamma_k} \\
        &=\left[dN-\sum_{k\in S}\gamma_k\right]\ln  \frac{\sigma^2(S\setminus\{j\})}{\sigma^2(S)} - \gamma_j \ln \frac{\lambda_j}{\gamma_j}  \\
        &\:\:\:\: +\gamma_j \ln \sigma^2(S\setminus\{j\}) \\
        &=\gamma_j \left( -\frac{dN - \sum_{k \in S} \gamma_k}{\gamma_j} \ln \frac{\sigma^2(S)}{\sigma^2(S\setminus\{j\})} - \ln \frac{\lambda_j}{\sigma^2(S\setminus \{j\}) \gamma_j} \right)
    \end{align*}
    }
Using \eqref{eq:sigma_new}, we find the following relation:
\begin{align*}
    \frac{\sigma^2(S)}{\sigma^2(S\setminus\{j\})} &= \frac{(dN - \sum_{k\in S} \gamma_k)/\gamma_j + 1 - \lambda_j/(\sigma^2(S\setminus\{j\})\gamma_j)}{(dN-\sum_{k\in S} \gamma_k)/\gamma_j}.
\end{align*}
To simplify notation, we introduce two quantities: 
\begin{align*}
    a := \frac{dN-\sum_{k\in S}\gamma_k}{\gamma_j} \geq 0,\quad&\quad 
    b := \frac{\lambda_j}{\gamma_j \sigma^2(S\setminus\{j\})} \geq 0,\\
    \textrm{leading to }\quad\quad \frac{\sigma^2(S)}{\sigma^2(S\setminus\{j\})} &= \frac{a + 1 - b}{a},\\
    \textrm{and }\quad\quad \frac{g(S\setminus\{j\}) - g(S)}{\gamma_j} &= -a \ln \frac{a+1-b}{a} - \ln b.
\end{align*}
We now show that the right-hand side quantity is positive.
Using $ 1+b \leq \exp(b)$, we find that $1+\ln(b) \leq b$. Thus,
\begin{align}
    \ln(b) &\leq b-1
    = -a \frac{1-b}{a}. \label{eq:klok_ineq1}
\end{align}
Using again for $z\geq -1$ that $\exp(z) \geq 1+z$, and so that $z \geq \ln(1+z)$ with $z = \frac{1-b}{a}$, we get: 
\begin{align*}
   \frac{1-b}{a} &\geq \ln \left(1 +\frac{1-b}{a}\right)   = \ln \left(\frac{a+1-b}{a}\right).
\end{align*}
\begin{align}
    &\textrm{Since $a\geq 0$, we get} -a \frac{1-b}{a} \leq -a \ln \left(\frac{a+1-b}{a}\right),\label{eq:klok_ineq2}\\
    &\textrm{and, combining \eqref{eq:klok_ineq1} and \eqref{eq:klok_ineq2},}\quad \ln(b) \leq -a \ln \left(\frac{a+1-b}{a}\right).\nonumber
\end{align}
This shows that $-a \ln \left(\frac{a+1-b}{a}\right) -\ln(b) \geq 0$ and, since $\gamma_j \geq 0$, that $\quad g(S\setminus\{j\})-g(S) \geq 0$.
\end{proof}

\begin{lemma}
\label{lem:saturated}
    There exists a solution $\hat{S}$ to \eqref{eq:optimization_problem} that meets the conditions of a \textbf{saturated} set: 
    \begin{align*}
    \hat{S} &\in \mathcal{V} &\text{and},\\
    \sigma^2(\hat{S}) &> \frac{\lambda_j}{\gamma_j}   &\forall j \in [K]\setminus \hat{S}.
\end{align*}
\end{lemma} 
\begin{proof}
    Since $\mathcal{V}$ is a finite set, there exists a solution to \eqref{eq:optimization_problem}.
    Let $\hat{S}$ be a solution to \eqref{eq:optimization_problem} that is not saturated.
    This means that $\hat{S} \in \mathcal{V}$, and there exist $j \in [K]\setminus \hat{S}$ such that
    \begin{align}
        \sigma^2(\hat{S}) &\leq \frac{\lambda_j}{\gamma_j}. 
        \label{eq:contradiction_cond}
    \end{align}
    From the definitions of $\sigma^2(\hat{S})$ and $\sigma^2(\hat{S}\cup\{j\})$ we know that 
    \begin{align*}
        \sigma^2(\hat{S}\cup\{j\}) &= \frac{\sigma^2(\hat{S})\left[dN - \sum_{k\in \hat{S}} \gamma_k\right] - \lambda_{j}}{dN - \sum_{k\in \hat{S}}\gamma_k - \gamma_{j}}.
    \end{align*}
    Rewriting \eqref{eq:contradiction_cond} as $\lambda_j \geq \sigma^2(\hat{S})\gamma_j$, and replacing it above, we find: 
    \begin{align*}
        \sigma^2(\hat{S}\cup\{j\}) 
        &\leq \frac{\sigma^2(\hat{S})\left[dN - \sum_{k\in \hat{S}} \gamma_k\right] - \sigma^2(\hat{S})\gamma_j}{dN - \sum_{k\in \hat{S}}\gamma_k - \gamma_j} =\sigma^2(\hat{S}).
    \end{align*}
    \begin{align*}
        \textrm{This shows that:}\quad\quad\quad \sigma^2(\hat{S}\cup\{j\}) &\leq \frac{\lambda_k}{\gamma_k} &\forall k\in \hat{S}, \\
        \sigma^2(\hat{S}\cup\{j\}) &\leq \frac{\lambda_j}{\gamma_j} &\text{from } \eqref{eq:contradiction_cond}.
    \end{align*}
    It thus follows from the definition of the valid sets $\mathcal{V}$ that $S \cup \{j\} \in \mathcal{V}$.
    Using Lemma~\ref{lem:decrease_set}, we find that $g(\hat{S}\cup\{j\}) \leq g(\hat{S})$. Moreover, since $\hat{S}$ is a solution to \eqref{eq:optimization_problem}, we must have
    \begin{align*}
        g(\hat{S}\cup\{j\}) &= g(\hat{S}),
    \end{align*}
    meaning that $\hat{S}\cup\{j\}$ is also a solution.
    We can recursively apply this reasoning on $\hat{S}\cup\{j\}$. Since $\hat{S} \subseteq [K]$, this recursion must end, at which point we get a saturated solution.    
\end{proof}

\begin{lemma}
\label{lem:unicity}
    There exists a unique \textbf{saturated} set.
\end{lemma}
\begin{proof}
    Suppose for the sake of contradiction that there exists two different saturated sets $S_1, S_2 \in \mathcal{V}$.
    $S_1$ being saturated, it satisfies the following inequalities: 
    \begin{align}
        \sigma^2(S_1) &\leq \frac{\lambda_k}{\gamma_k} &\forall k \in S_1,\label{eq:s1_cond1}\\
        \sigma^2(S_1) &> \frac{\lambda_k}{\gamma_k} &\forall k\in [K]\setminus S_1 \label{eq:s1_cond2}.
    \end{align}
    Similarly, for $S_2$ we have: 
    \begin{align}
        \sigma^2(S_2) &\leq \frac{\lambda_k}{\gamma_k} &\forall k \in S_2, \label{eq:s2_cond1}\\
        \sigma^2(S_2) &> \frac{\lambda_k}{\gamma_k} &\forall k\in [K]\setminus S_2. \label{eq:s2_cond2}
    \end{align}
    \begin{align}
        & \textrm{Without loss of generality, assume  }\sigma^2(S_1)\leq \sigma^2(S_2).
        \label{eq:relation_s1_2}
    \end{align}
    Combining \eqref{eq:relation_s1_2} with \eqref{eq:s2_cond1}, we find:
    \begin{align*}
        \sigma^2(S_1) &\leq \frac{\lambda_k}{\gamma_k} &\forall k \in S_2.
    \end{align*}
    Using \eqref{eq:s1_cond2}, this means that : $\forall j \in S_2 \: : 
        \: j\notin [K]\setminus S_1,$
    \noindent
    or equivalently, that $S_2 \subset S_1$ (the strict inclusion comes from the assumption $S_1 \neq S_2$).
\noindent    
We deduce the following relation: 
    \begin{align}
        \sigma^2(S_1) &= \frac{\sigma^2(S_2)\left[dN - \sum_{k\in S_2} \gamma_k\right] - \sum_{j\in S_1\setminus S_2} \lambda_j}{dN - \sum_{k\in S_2} \gamma_k - \sum_{j\in S_1\setminus S_2} \gamma_j} 
        \label{eq:sigma_1_2}
    \end{align}
From \eqref{eq:s2_cond2}, we know: $\lambda_j < \gamma_j \sigma^2(S_2)$ for $j \in S_1\setminus S_2$. Replacing this in \eqref{eq:sigma_1_2}, we find
\begin{align*}
    \sigma^2(S_1) &> \frac{\sigma^2(S_2)\left[dN - \sum_{k\in S_2} \gamma_k\right] - \sigma^2(S_2) \sum_{j\in S_1\setminus S_2} \gamma_j}{dN - \sum_{k\in S_2} \gamma_k - \sum_{j\in S_1\setminus S_2} \gamma_j}  \\
    &= \sigma^2(S_2).
\end{align*}
This contradicts \eqref{eq:relation_s1_2}.
\end{proof}
\noindent
From Lemma~\ref{lem:saturated}, we know that there exists a solution to \eqref{eq:optimization_problem} that is a \textbf{saturated} set. Moreover, we know from Lemma~\ref{lem:unicity} that the saturated set is unique. This means that the saturated set is a solution to \eqref{eq:optimization_problem}.
Algorithm~\ref{alg:greedy_set} is a greedy algorithm finding that \textbf{saturated} solution in $\mathcal{O}(K^2)$.

\begin{algorithm}[H]
\caption{Greedy set optimization for \eqref{eq:optimization_problem}}\label{alg:greedy_set}
\begin{algorithmic}
\STATE
\STATE {\textsc{Input:}}
\STATE \hspace{0.5cm} $\lambda_1, \ldots , \lambda_K$,$\gamma_1 , \ldots , \gamma_K$, $\|Y\|_\mathrm{F}^2$
\STATE {\textsc{Initialization :}}
\STATE \hspace{0.5cm} $S = \emptyset$
\STATE \hspace{0.5cm} $L = [K]$
\STATE {\textsc{While } $L \neq \emptyset$:}
\STATE \hspace{0.5cm} $\sigma^2(S) = \left[ \|\Y\|_\mathrm{F}^2 - \sum_{k\in S}^{K} \lambda_k \right]/(dN - \sum_{k \in S} \gamma_k) $
\STATE \hspace{0.5cm} $L = \left\{k\in [K]\setminus S \:\: \text{s.t} \:\: \sigma^2(S)\leq \lambda_k/\gamma_k \right\}$
\STATE \hspace{0.5cm} Choose any $k^\star \in L$
\STATE \hspace{0.5cm} $S = S \cup \{k^\star\}$
\STATE {\textsc{Output:}}
\STATE \hspace{0.5cm} $S$
\end{algorithmic}
\end{algorithm}
\noindent
At every step of algorithm~\ref{alg:greedy_set}, the set $S$ is a valid set. This can be seen by induction. For $S\in \mathcal{V}$, and $k^\star \in L$, we have: 
\begin{align}
    \sigma^2(S)\leq \lambda_{k^\star}/\gamma_{k^\star},\quad\quad\textrm{ and thus,}
    \label{eq:sigma_k_star}
\end{align}
\begin{align*}
    \sigma^2(S\cup\{k^\star\}) 
    &= \frac{\sigma^2(S)\left[dN-\sum_{k\in S} \gamma_k \right] - \lambda_{k^\star}}{dN - \sum_{k\in S}\gamma_k - \gamma_{k^\star}} \\
    &\leq \frac{\sigma^2(S)\left[dN-\sum_{k\in S} \gamma_k \right] - \sigma^2(S)\gamma_{k^\star}}{dN - \sum_{k\in S}\gamma_k - \gamma_{k^\star}} =\sigma^2(S).
\end{align*}
Using that $S \in \mathcal{V}$, and \eqref{eq:sigma_k_star} we find: 
\begin{align*}
    \sigma^2(S\cup\{k^\star\}) &\leq \frac{\lambda_j}{\gamma_j} &&\forall j \in S, \textrm{ and}\\
    \sigma^2(S\cup\{k^\star\}) &\leq \frac{\lambda_{k^\star}}{\gamma_{k^\star}}, &&\textrm{showing that $S\cup\{k^\star\}$ is also in $\mathcal{V}$.}
\end{align*}
\noindent
Algorithm~\ref{alg:greedy_set} must terminate since set $L$ decreases at every step. When it terminates, set $S$ is \textbf{saturated} since $L$ is empty.

\section{Proof of solution to \ac{GMM} retrieval}
\label{app:gmm_proof}

\begin{lemma}
    \label{lem:rank1_projection}
    For $\A_1, \ldots \A_K$, $K$ symmetric matrices in $\mathbb{R}^{d\times d}$, and $\x_1 , \ldots \x_K \in \mathbb{R}^d$, we have the following:
    \begin{align}
    \sum_{k=1}^K \|\A_k - \x_k \x_k ^\mathrm{T}\|_\mathrm{F}^2 &\geq \sum_{k=1}^K \sum_{i=2}^d \lambda_i^2(\A_k), \label{eq:GMM_ineq}
    \end{align}
    where $\lambda_1(\A_k)$ is the leading eigenvalue of $\A_k$.
    With $\vu_k$ the eigenvector associated to $\lambda_1(\A_k)$, \eqref{eq:GMM_ineq} holds with equality whenever, for all $k$: $\lambda_1(\A_k) \geq 0$ and $\x_k = \sqrt{\lambda_1(\A_k)} \vu_k$.
\end{lemma}
\begin{proof}
Expanding the negative of the left-hand side, we find:
    \begin{align*}
        -\sum_{k=1}^K \|\A_k - \x_k \x_k ^\mathrm{T}\|_\mathrm{F}^2 = \sum_{k=1}^K -\textrm{tr}(\A_k^2)+2 \x_k^\mathrm{T} \A_k \x_k  - \|\x_k\|^4.
    \end{align*}
    Combining this with $\x_k^\mathrm{T} \A_k \x_k \leq \lambda_1(\A_k)\|\x_k\|^2$ for $\lambda_1(\A_k)$, the leading eigenvalue of $A_k$, we find : 
    {\allowdisplaybreaks
    \begin{align}
    &-\sum_{k=1}^K \|\A_k - \x_k \x_k ^\mathrm{T}\|_\mathrm{F}^2 \nonumber \\ 
    &\leq \sum_{k=1}^K \left[2 \lambda_1(\A_k) \|\x_k\|^2 - \|\x_k\|^4 -\textrm{tr}(\A_k^2)\right] \label{eq:gmm_ineq1}\\
    &= \sum_{k=1}^K \left[-(\lambda_1(\A_k) - \|\x_k\|^2)^2 + \lambda_1^2(\A_k) - \textrm{tr}(\A_k^2)\right] \nonumber \\
    &\leq \sum_{k=1}^K \left[\lambda_1^2(\A_k) -\textrm{tr}(\A_k^2)\right] = -\sum_{k=1}^K \sum_{i=2}^d \lambda_i^2(\A_k). \label{eq:gmm_ineq2}
    \end{align}
    }
    Note that \eqref{eq:gmm_ineq1} holds with equality whenever $\x_k$ is a rescaled eigenvector $\A_k$ with eigenvalue $\lambda_1(\A_k)$. Also, the inequality in \eqref{eq:gmm_ineq2} can only hold with equality whenever $\lambda_1(\A_k) \geq 0$, and $\|\x_k\|^2 = \lambda_1(\A_k)$.
\end{proof}
\begin{lemma}
    \label{lem:minimizer_gmm}
    Let $\hat{\Sig}_1, \ldots , \hat{\Sig}_K \in \mathbb{R}^{d\times d}$ be symmetric positive semi-definite matrices, such that
    \begin{align}
        \lambda_1(\hat{\Sig}_j) &\geq \frac{\sum_{k=1}^K \sum_{i=2}^d \lambda_i(\hat{\Sig}_k)}{K(d-1)} &\forall j \in [K], \label{eq:extra_condition_gmm}
    \end{align}
    where $\lambda_i(\hat{\Sig}_k)$ is the $i$-th largest singular value of $\hat{\Sig}_k$. The solution to 
    \begin{equation}
    \begin{array}{lcl}
        &\min & \sum_{k=1}^K \|\hat{\Sig}_k - \left(\hat{\x}_k\hat{\x}_k^\mathrm{T}+ \hat{\sigma}^2 I\right)\|_\mathrm{F}^2 \\
&\hat{\x}_1,\ldots,\hat{\x}_K,\hat{\sigma}^2
    \end{array},
    \end{equation}
    \begin{align*}
        \textrm{then is }\quad\sigma^2_{\text{GMM}} &= \frac{1}{K}\sum_{k=1}^K \frac{\textrm{tr}(\hat{\Sig}_k) - \lambda_1(\hat{\Sig}_k)}{d-1}, \\
         \x_{\text{GMM},k} &= \sqrt{\left(\lambda_1(\hat{\Sig}_k)-\sigma^2_{\text{GMM}}\right)} \: \vb_1(\hat{\Sig}_k) &\forall k \in [K],
    \end{align*}
    where $\vb_1(\hat{\Sig}_k)$ is an eigenvector associated to $\lambda_1(\hat{\Sig}_k)$
\end{lemma}
\begin{proof}
    Using Lemma~\ref{lem:rank1_projection} for $\A_k := \hat{\Sig}_k - \sigma^2 I $, we find
    \begin{align}
    \sum_{k=1}^K \|\hat{\Sig}_k - \left(\hat{\x}_k\hat{\x}_k^\mathrm{T}+ \sigma^2 I\right)\|_\mathrm{F}^2 &\geq \sum_{k=1}^K \sum_{i=2}^d \lambda_i^2(\hat{\Sig}_k - \sigma^2 I) \label{eq:tight_gmm} \\
    &= \sum_{k=1}^K \sum_{i=2}^d (\lambda_i(\hat{\Sig}_k) - \sigma^2)^2. \nonumber
    \end{align}
    We find the $\sigma^2$ that minimizes the previous equation by setting the derivative to zero. This leads us to 
    \begin{align*}
        \sigma_{GMM}^2 &= \frac{\sum_{k=1}^K \sum_{i=2}^d \lambda_i(\hat{\Sig}_k)}{K(d-1)} = \frac{1}{K}\sum_{k=1}^K\frac{\textrm{tr}(\hat{\Sig}_k) - \lambda_1(\hat{\Sig}_k)}{d-1}.
    \end{align*}
    \begin{align*}
        \textrm{Condition \eqref{eq:extra_condition_gmm} implies that}\quad &\sigma_{GMM}^2 \leq \lambda_1(\hat{\Sig}_k) &\forall k \in [K],\\
        \textrm{and thus that}\quad &\lambda_1(\hat{\Sig}_k - \sigma^2_{\text{GMM}} I) \geq 0 &\forall k \in [K].
    \end{align*}
    Finally, we know from Lemma~\ref{lem:rank1_projection} that \eqref{eq:tight_gmm} is tight when 
    \begin{align*}
        \hat{\x}_k &= \x_{\text{GMM},k} = \sqrt{\lambda_1(\hat{\Sig}_k)-\sigma^2_{\text{GMM}}} \: \vb_1(\hat{\Sig}_k) &\forall k \in [K].
    \end{align*}
\end{proof}

\end{appendices}
\vspace*{-0.4cm}
\bibliographystyle{IEEEtranN}
\bibliography{reference}


\begin{IEEEbiography}[{\includegraphics[width=1.1in,height=1.2in,clip,keepaspectratio]{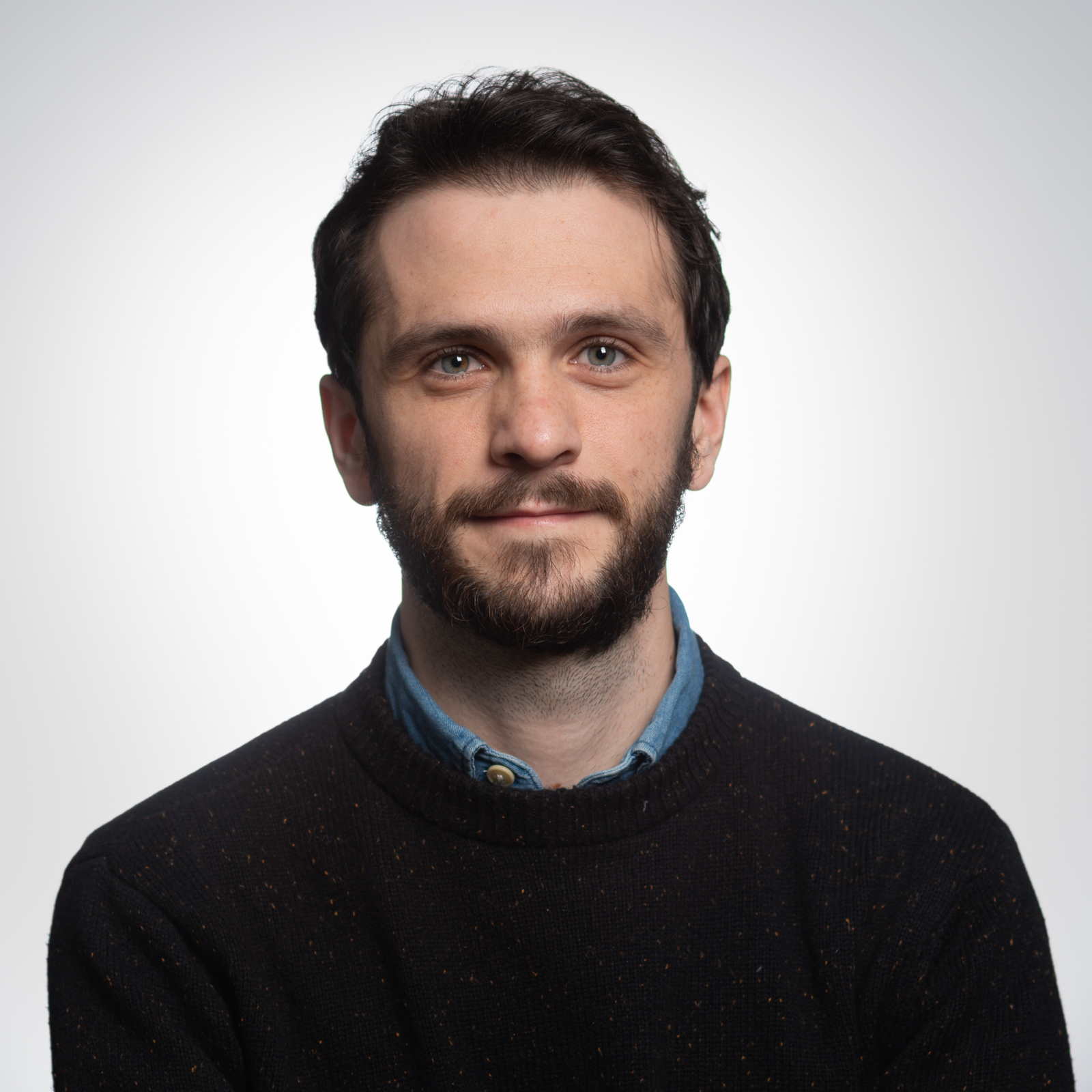}}]{Paul-Louis Delacour}
received the B.Sc. degree in communication systems (computer science) from École Polytechnique Fédérale de Lausanne (EPFL), Lausanne, Switzerland, in 2019 and the M.Sc. degree in data science from the Federal Institute of Technology Zurich (ETH Zurich), Zurich, Switzerland, in 2022. He is currently working towards the Ph.D. degree at Delft University of Technology (TU Delft), Delft, The Netherlands.
His research interest includes mathematics \& machine learning with a focus on high-dimensional learning.
\end{IEEEbiography}

\begin{IEEEbiography}[
{\includegraphics[width=1in,height=1.25in,clip,keepaspectratio]{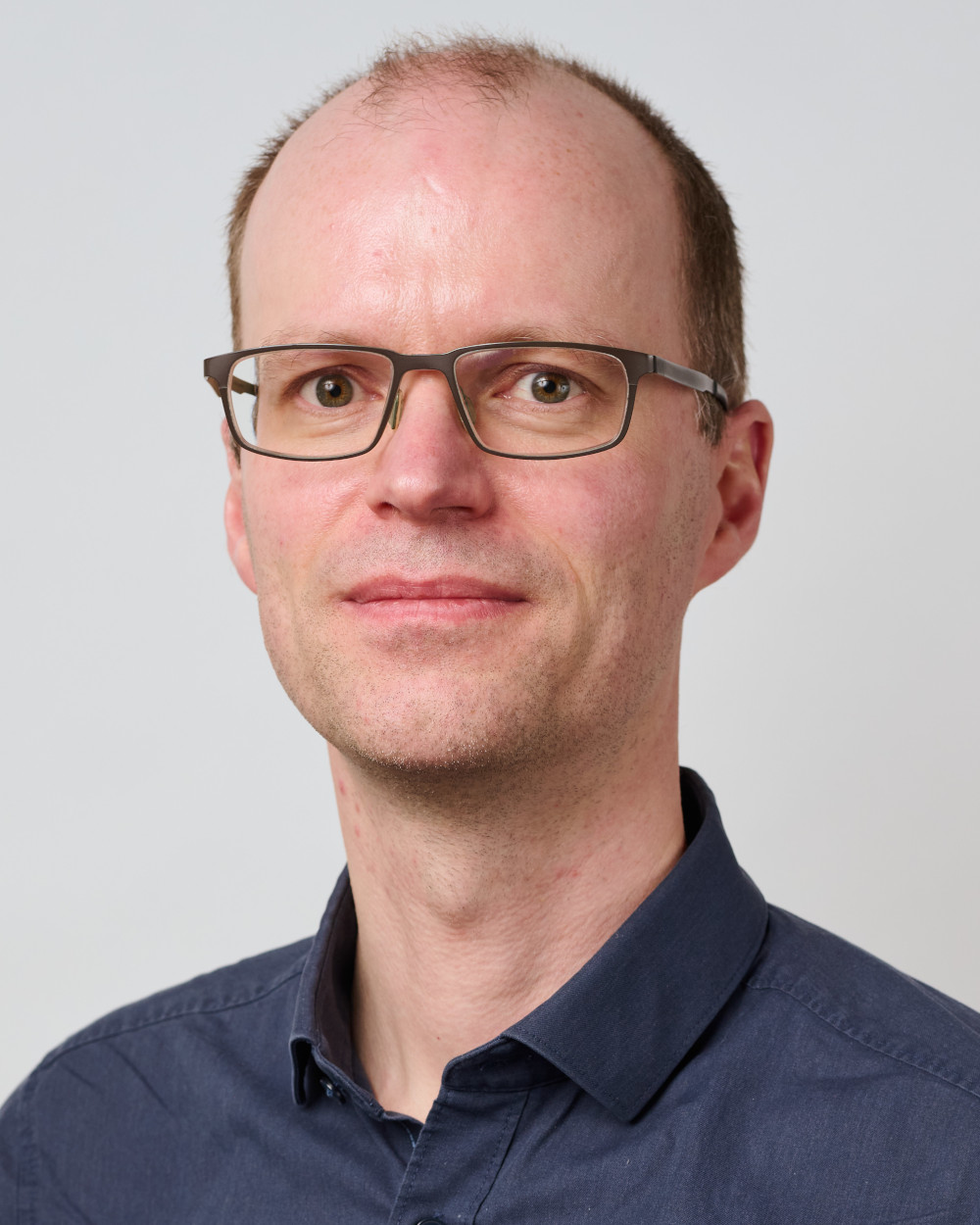}}
]
{Sander Wahls (S'09--M'12--SM'17)}
received the Dipl.-Math. degree in mathematics and the Dr.-Ing. degree (summa cum laude) in electrical engineering from Technische Universität Berlin, Berlin, Germany, in 2007 and 2011, respectively.

He is currently a Full Professor of information technology at Karlsruhe Institute of Technology, Karlsruhe, Germany. From 2014 to 2023, he was with Delft University of Technology, Delft, The Netherlands, first as Assistant Professor, and later as Associate Professor. From 2012 to 2014, he was a Postdoctoral Research Fellow at Princeton University, Princeton, NJ, USA.

Dr. Wahls is a member of VDE-ITG and SIAM. He received the Johann-Philipp-Reis Preis from VDE-ITG in 2015, and a Starting Grant from the European Research Council (ERC) in 2016.
\end{IEEEbiography}

\begin{IEEEbiography}[{\includegraphics[width=1in,height=1.25in,clip,keepaspectratio]{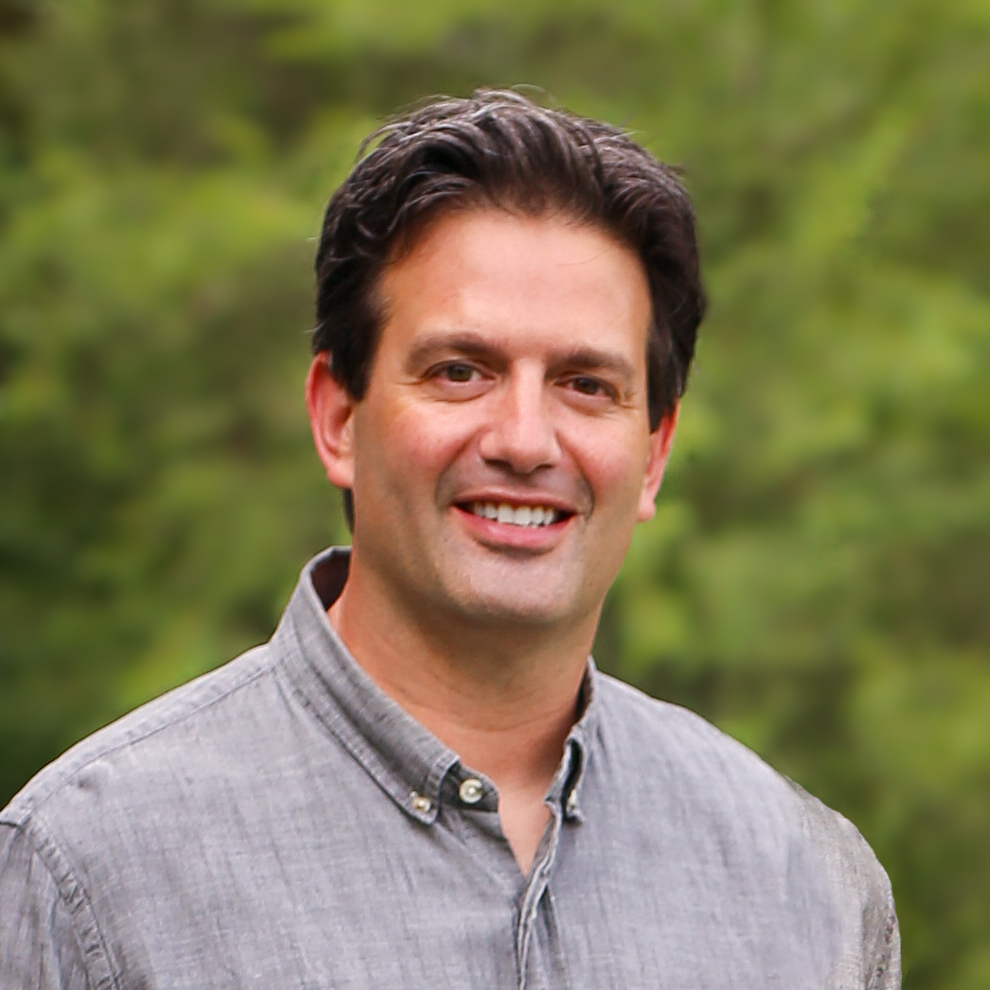}}]{Jeffrey M. Spraggins}
received the B.A. degree in chemistry from the College of Wooster and the Ph.D. degree in analytical chemistry from the University of Delaware in 2010.

He is currently Associate Professor in the Department of Cell and Developmental Biology and Director of both the Mass Spectrometry Research Center (MSRC) and Biomolecular Multimodal Imaging Center (BIOMIC) at Vanderbilt University, Nashville, TN, USA.
Dr. Spraggins’ research focuses on two main areas: (1) the development of advanced mass spectrometry technologies to enhance imaging performance and enable molecular histology at cellular resolution, and (2) combining imaging mass spectrometry with other complementary biomedical imaging modalities to create new, integrated platforms for systems-level analysis of tissue biology. His group applies these approaches to the construction of comprehensive molecular atlases for the NIH Human Biomolecular Atlas Program (HuBMAP), the NIDDK Kidney Precision Medicine Project (KPMP), and the NCI Human Tumor Atlas Network (HTAN). In parallel, his team investigates the molecular underpinnings of Alzheimer’s, kidney, and infectious diseases.
\end{IEEEbiography}

\begin{IEEEbiography}[{\includegraphics[width=1in,height=1.25in,clip,keepaspectratio]{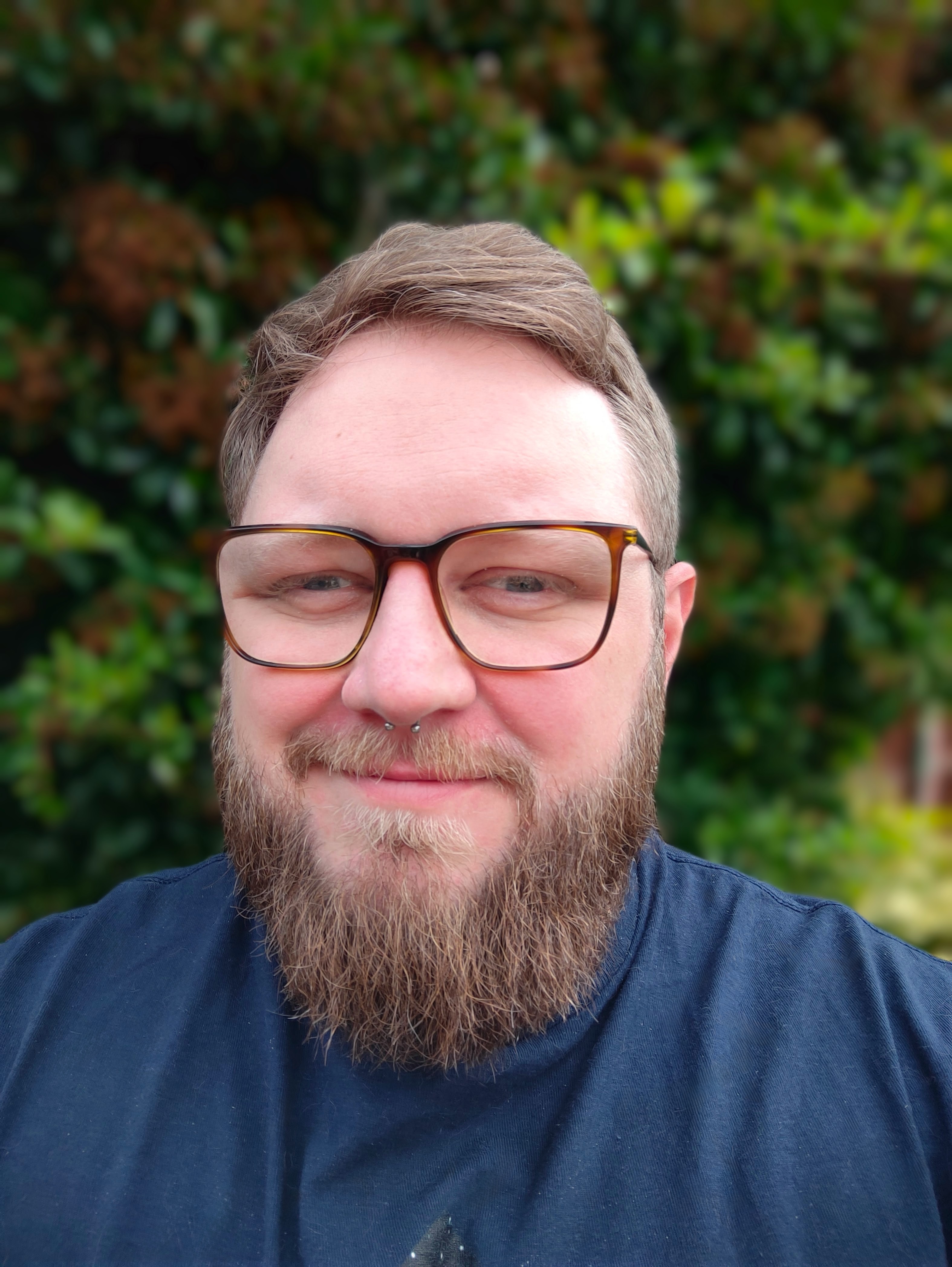}}]{Lukasz Migas}
received the MChem degree in chemistry from Manchester Metropolitan University, Manchester, UK, in 2014, and the Ph.D. degree in biochemistry from the University of Manchester, Manchester, UK, in 2018. He is currently a Postdoctoral Research Fellow at the Delft Center for Systems and Control, Delft University of Technology, Delft, The Netherlands, and at the Mass Spectrometry Research Center, Vanderbilt University, Nashville, TN, USA. His research interests include the development of computational methods and user-friendly graphical applications for the analysis and interpretation of imaging mass spectrometry and microscopy data.
\end{IEEEbiography}

\begin{IEEEbiography}[{\includegraphics[width=1in,height=1.25in,clip,keepaspectratio]{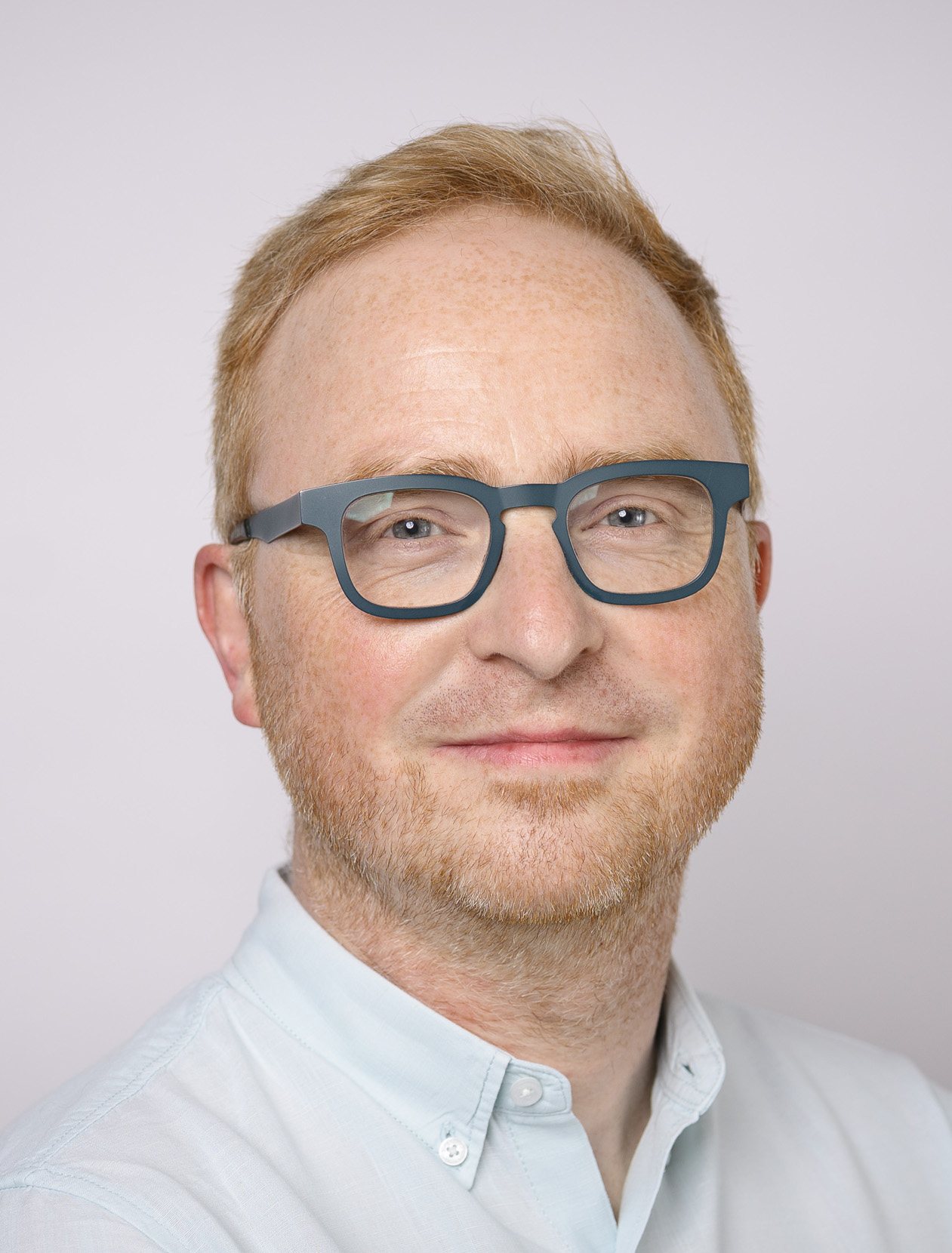}}]{Raf Van de Plas}
(Member, IEEE) received the Industrieel Ingenieur degree in electronics from Groep T, Leuven, Belgium, in 2002. He received the Master in Artificial Intelligence degree and the Doctor of Engineering (Ph.D.) degree from the Katholieke Universiteit Leuven, Leuven, Belgium, in 2003 and 2010, respectively.

He was a Postdoctoral Researcher in the Department of Electrical Engineering (ESAT) at the Katholieke Universiteit Leuven in 2010, and a Research Fellow in the Department of Biochemistry through the Mass Spectrometry Research Center at Vanderbilt University, Nashville, TN, USA, from 2011 to 2012. During 2011, he was an Honorary Fulbright Scholar at Vanderbilt University. From 2012 to 2014, he joined Vanderbilt University's School of Medicine research faculty as Research Instructor. From 2014 to 2022, he was Assistant Professor in the Delft Center for Systems and Control at the Delft University of Technology (TU Delft), Delft, The Netherlands. Since 2014, he holds a position as Adjunct/Adjoint Assistant Professor of Biochemistry at the Vanderbilt University School of Medicine. He is currently Associate Professor in the Delft Center for Systems and Control at the Delft University of Technology. He works on signal processing and machine learning for spectral imaging systems (\textit{e.g.}, imaging mass spectrometry), molecular imaging modalities (\textit{e.g.}, microscopy), and data-driven multimodal fusion between imaging technologies. His research focuses on the interface between (i) mathematical engineering and machine learning; (ii) analytical chemistry and instrumentation; and (iii) life sciences and medicine.

Dr. ing. Van de Plas is a member of the American Society for Mass Spectrometry (ASMS), the Belgian Society for Mass Spectrometry (BSMS), and the Dutch Society for Mass Spectrometry (NVMS).\end{IEEEbiography}

\newpage
\clearpage
\manualtitle
\appendix
\section*{Supplementary Material}
\subsection{Proof of Lemma~\ref{lem:approximated_Ak}}
\label{sec:proof_results}
\begin{proof}
    \begin{align*}
        &|\lambda_1(\A_k) - \lambda_1(\Tilde{\A}_k)| \\
        &\leq \norm{  \sum_{\substack{i=1 \\ \rho_k^{[i]} \leq \tau}}^N \rho_k^{[i]}\y_i \y_i^\mathrm{T}  }_{op} &\text{Weyl's inequality} \\
        &=\lambda_1\left(\sum_{\substack{i=1 \\ \rho_k^{[i]} \leq \tau}}^N \rho_k^{[i]}\y_i\y_i^\mathrm{T}\right) &\text{psd matrix} \\
        &\leq \tau \lambda_1\left(\sum_{\substack{i=1 \\ \rho_k^{[i]} \leq \tau}}^N \y_i\y_i^\mathrm{T}\right) &\text{Increase eigenvalues} \\
        &\leq \tau \sum_{\substack{i=1 \\ \rho_k^{[i]} \leq \tau}}^N \|\y_i\|^2 &\text{$\textrm{tr}(.)\geq \lambda_1(.)$ for psd matrix} \\
        &= \delta\frac{1}{\|\Y\|_\mathrm{F}^2} \sum_{\substack{i=1 \\ \rho_k^{[i]} \leq \tau}}^N \|\y_i\|^2 &\text{definition of }\tau \\
        &\leq \delta &\sum_{\substack{i=1 \\ \rho_k^{[i]} \leq \tau}}^N \|\y_i\|^2 \leq \|\Y\|_\mathrm{F}^2
    \end{align*}
\end{proof}

\subsection{Hyperparameters specifications}
In many iterative algorithms, sieving is a popular technique to improve efficiency by removing runs that appear irrelevant or suboptimal as the algorithm progresses.
In the context of \ac{EM}, this consists of repeating the procedure multiple times for different starting points, finding a starting point's maximum likelihood each time, spending more resources only on the best candidates, and finally retaining the best one.
More precisely, in our \ac{SMM}-fitting cases, we carry out a two-stage approach in which we first perform $d_1$ (pre-sieving) iterations of \ac{EM} for $n_1$ different random initializations.
Of those $n_1$ pre-sieving starting points, we select (or `sieve') $n_2$ of the best performing initializations, on which we then perform $d_2$ (post-sieving) iterations of \ac{EM}.
Furthermore, we stop iterating when the likelihood gain is below a certain threshold $t$.
For the synthetic, \ac{HSI}, and \ac{IMS} datasets, the values of $n_1, d_1 , n_2 , d_2, t$ is given in Table~\ref{tab:smm_parameters}.

\begin{table}[t]
    \centering
    \begin{tabular}{|r||c|c|c|c|}
        \hline
        & Synthetic & Rat brain \ac{IMS} & Salient \ac{HSI} \\
        \hline
        starting points $n_1$ & $10$ & $6$ & $10$ \\
        \hline 
         pre-sieving iterations $d_1$ & $10$ & $6$ & $10$ \\
         \hline 
         selected points $n_2$ & $5$ & $3$ & $3$ \\
         \hline
         post-sieving iterations $d_2$ & $600$ & $60$ & $120$ \\
         \hline
         early stop gain $t$ & $10^{-8}$ & $10^{-3}$ & $10^{-3}$ \\
         \hline
    \end{tabular}
    \newline
    \caption{\ac{SMM} hyperparameter specifications}
    \label{tab:smm_parameters}
\end{table}

\subsection{
\textcolor{manualcolor}{
\ac{HSI} dataset; impact of scaling on the recovered signals
}
}
\begin{figure*}[ht]
\centering
\includegraphics[width=0.6\textwidth]{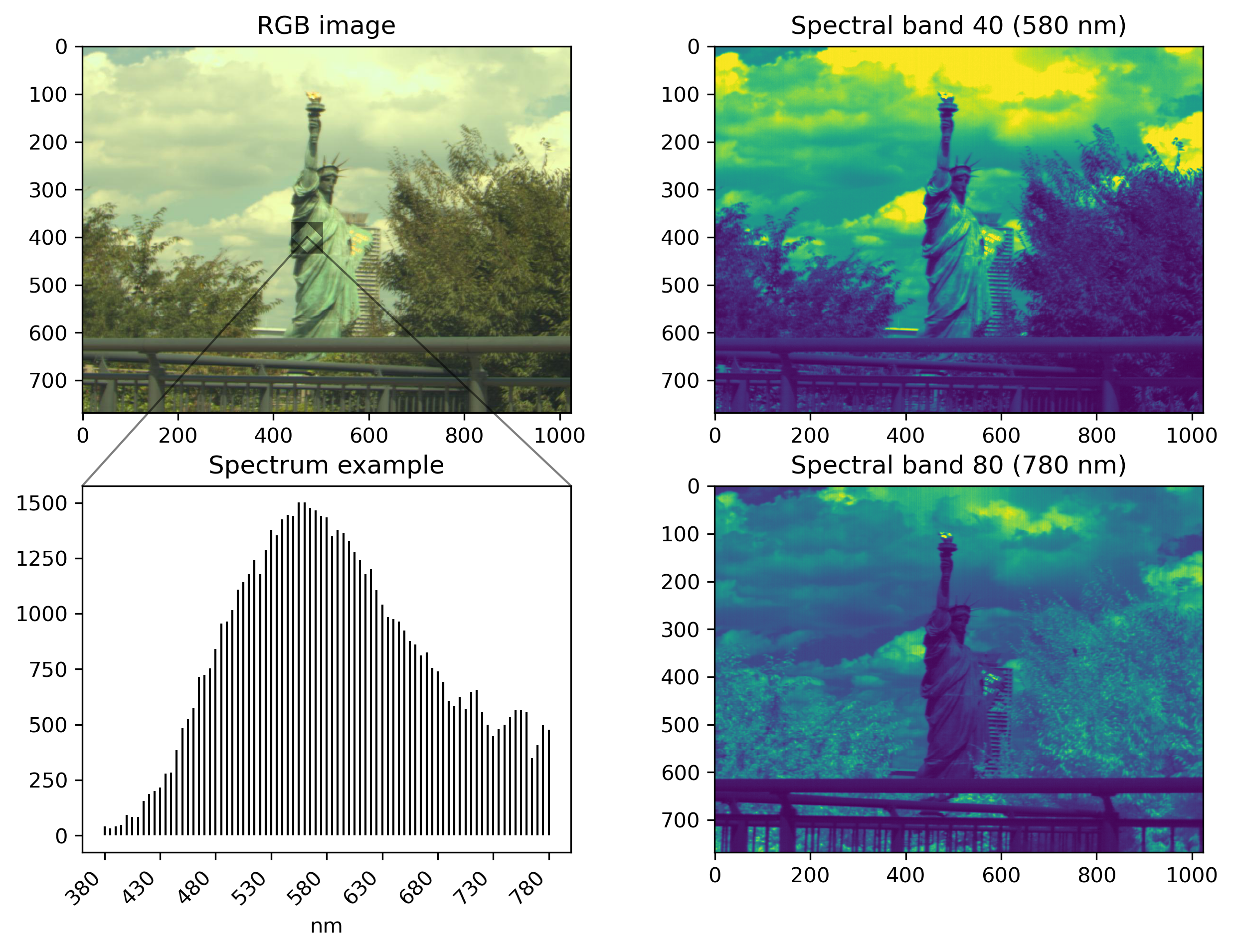}
\caption{Salient \ac{HSI} dataset. (top-left) A RGB reference image covering the scene measured by the Salient \ac{HSI} dataset from \cite{salient_data}. One can recognize the Statue of Liberty, a building behind it, and foliage and a railing structure in the front. (right) Two of the
81 wavelength-specific images acquired, showing the spatial distributions of hyperspectral bands $580$ nm and $780$ nm. (bottom-left) An example electromagnetic spectrum acquired at a single pixel, reporting 81 distinct wavelength features.}
\label{fig:salient_data}
\end{figure*}

The \ac{HSI} dataset captures both spatial and spectral information on the measured scene. Figure~\ref{fig:salient_data} illustrates the \ac{HSI} dataset from~\cite{salient_data}.

It should be noted that the type of data normalization that is applied to a dataset prior to signal estimation and recovery can substantially impact the final result.
For example, a feature-wise normalization will scale every feature in the dataset separately, while an observation-specific normalization affects an entire observation, but will often keep inter-feature relationships within an observation intact.
Although data normalization and scaling are in essence separate from the signal recovery step, it is clear that the signals recovered will differ for different scalings applied.
It is therefore important to consider whether normalization should be applied prior to signal recovery and if so, which type of normalization is most appropriate.
In many cases, this is a consideration specific to the application domain at hand.

In the case studies described in the main paper, we employ min-max scaling prior to conducting signal recovery.
Min-max scaling is a data normalization that operates in a feature-wise manner, mapping all features to the range $[0,1]$.
For every feature $f$, the rescaled observation $y_i^\prime$ is linked to the original observation $y_i$ as follows:
\begin{align*}
    y_i^\prime[f]={\frac {y_i[f]-\min_j(y_j[f])}{\max_{j}(y_j[f])-\min_{j}(y_j[f])}}.
\end{align*}

To demonstrate the impact of a chosen data normalization on signal recovery, we also explore another normalization, namely observation-wide $l_1$-norm-based scaling.
The $l_1$-norm-based approach acts on every observation separately.
It does not treat the feature values within an observation separately, but instead rescales an entire observation $y_i^\prime$ as follows: 
\begin{align*}
    y_i^\prime = \frac{y_i}{\|y_i\|_1}.
\end{align*}
Thus, unlike min-max scaling, with $l_1$-norm-based normalization all feature values within an observation receive the same scaling factor.

\begin{figure*}[ht]
\centering
\begin{minipage}[b]{0.5\textwidth}
    \subfloat[Min-max normalization]{\includegraphics[width=\textwidth]{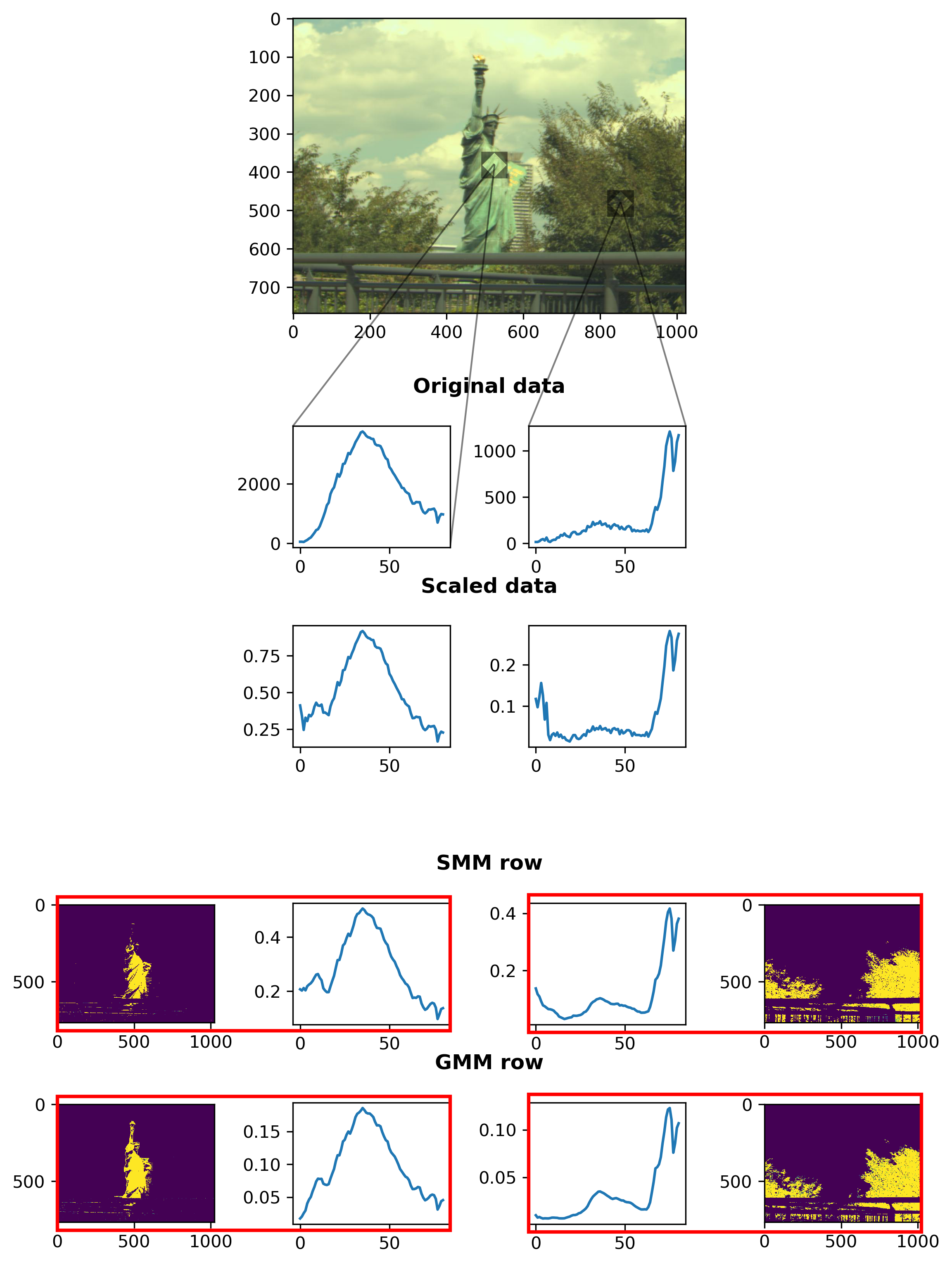}\label{fig:min_max_salient_min_max}
    }
\end{minipage}%
\begin{minipage}[b]{0.5\textwidth}
    \subfloat[l$1$ normalization]
{\includegraphics[width=\textwidth]{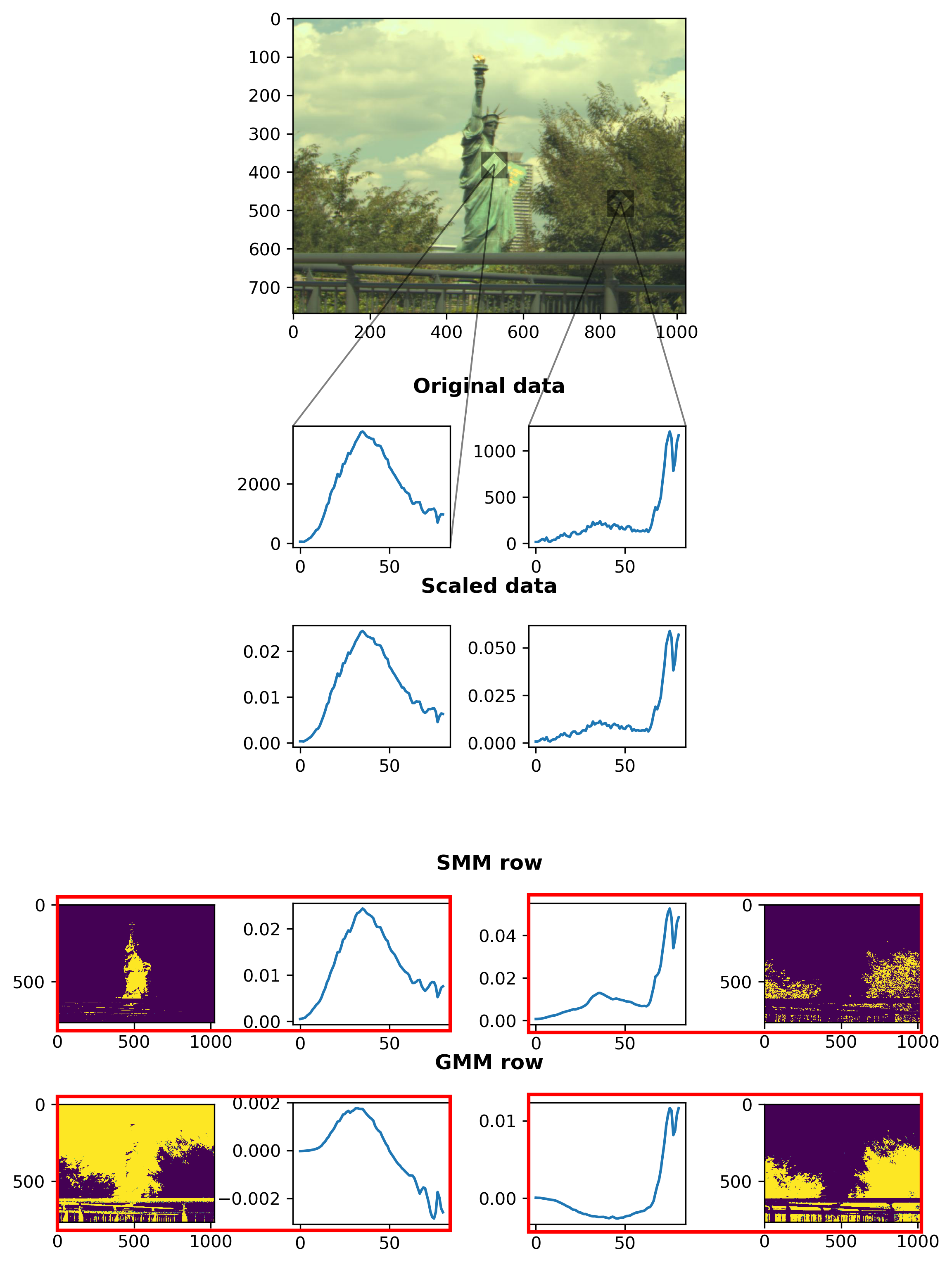}\label{fig:l1_saleint}
    \label{fig:min_max_salient_l1}
}
\end{minipage}%
\caption{Comparison of estimated spectra recovered from the \ac{HSI} dataset for two different prior data normalizations. (left) Figure~\ref {fig:min_max_salient_min_max} shows the scaled observations $\y_i$ and recovered signals $\x_i$, after applying min-max normalization. The first row below the RGB image shows two unnormalized example spectra from the original dataset, and the row below depicts their rescaled version. The last two rows show the associated estimated spectra obtained by \ac{SMM} and \ac{GMM} fitting. (right) Figure~\ref{fig:min_max_salient_l1} shows the same analysis, albeit using $l_1$-norm-based normalization instead. Notice the difference in recovered relative signal profiles between the two normalization approaches, and the substantial presence of negative values in the signals recovered by \ac{GMM} after $l_1$-norm-based scaling (despite the original data being non-negative). It is clear that data normalization prior to signal recovery is an important consideration to make before performing a mixture model fitting.}
\label{fig:normalization_salient}
\end{figure*}

Figure~\ref{fig:normalization_salient} shows the results of \ac{SMM} and \ac{GMM} fitting by \ac{EM} on the \ac{HSI} dataset after applying the two normalizations described above. 
This figure shows, below the RGB image, two example spectra and their rescaled versions.
The spectra after min-max normalization are shown on the left in Fig.~\ref{fig:min_max_salient_min_max} and the same spectra after $l_1$-norm-based scaling are shown on the right in Fig.~\ref{fig:min_max_salient_l1}.
Since the $l_1$-norm normalization is observation-specific, the scaled spectra in Fig.~\ref{fig:min_max_salient_l1} exhibit the same relative spectral profile, only scaled onto a different absolute value range.
The spectra in Fig.~\ref{fig:min_max_salient_min_max} show changes in their relative spectral profile post-scaling due to min-max normalization treating each feature separately.
The two bottom rows in Fig.~\ref{fig:normalization_salient} provide signal recovery results after applying \ac{SMM} and \ac{GMM} fitting by \ac{EM}.
The red boxes show two signal subpopulations recovered per normalization-mixture model combination.
Although it is difficult to assess without external information, and as far as one can tell by visual inspection, the signals estimated by \ac{SMM} seem more accurate in both normalization cases.
Furthermore, the effect of normalization on a signal recovery run is clearly visible in the bottom-right \ac{GMM} case with $l_1$-norm-based scaling, where the estimated signals contain negative values while the original data do not.

\begin{figure*}[ht]
    \centering
    \includegraphics[width=0.90\linewidth]{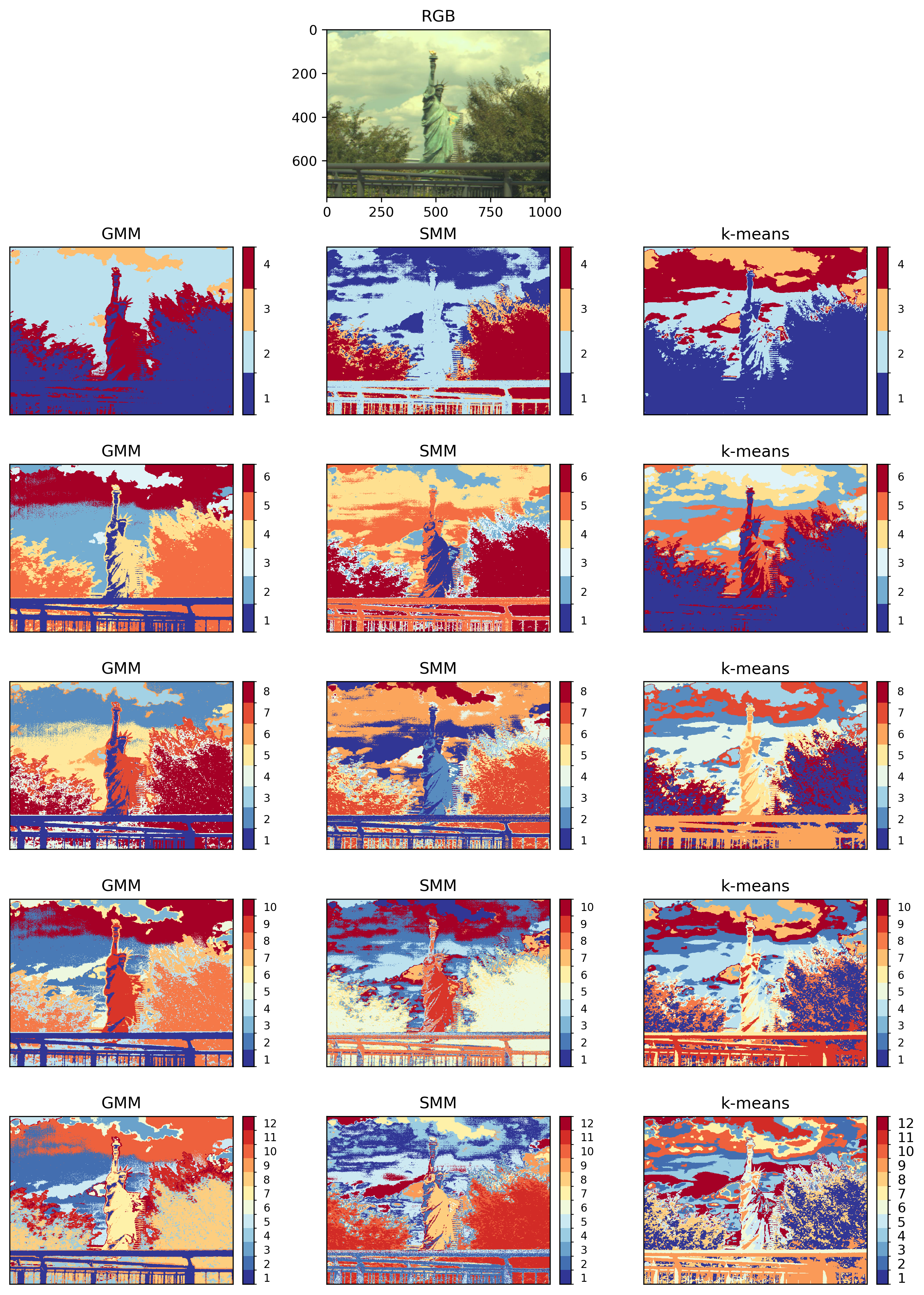}
    \caption{Comparison of the clustering results for the \ac{HSI} dataset as obtained by \ac{SMM} fitting by \ac{EM}, \ac{GMM} fitting by \ac{EM}, and \acl{KMC}, for different values of $K \in [4,6,8,10,12]$ and using min-max normalization.}
    \label{fig:all_k_salient}
\end{figure*}

For completeness, in Figure~\ref{fig:all_k_salient} we provide for the \ac{HSI} dataset the clustering results employing \ac{SMM}, \ac{GMM}, and \ac{KMC} for $K \in [4,6,8,10,12]$. 
\textcolor{manualcolor}{
In practice, we could estimate the number of clusters $K$ using metrics such as the Silhouette score, assessing cluster compactness and separation, as well as the \ac{AIC} or the \ac{BIC}. Alternatively, we could use a cross-validation approach, splitting the dataset into training and validation sets, then evaluating the likelihood on the validation set for various values of $K$. For the case study in the paper, we manually chose a value of $K=10$, to highlight the fact that there are real-world scenarios in which our \ac{SMM} approach is able to find separation that \ac{GMM} and \textit{k}-means omit, as mentioned in Figure~$8$. 
}

\subsection{Rat brain \ac{IMS} dataset and recovered signals}
\begin{figure*}[ht]
    \centering
    \includegraphics[width=1\linewidth]{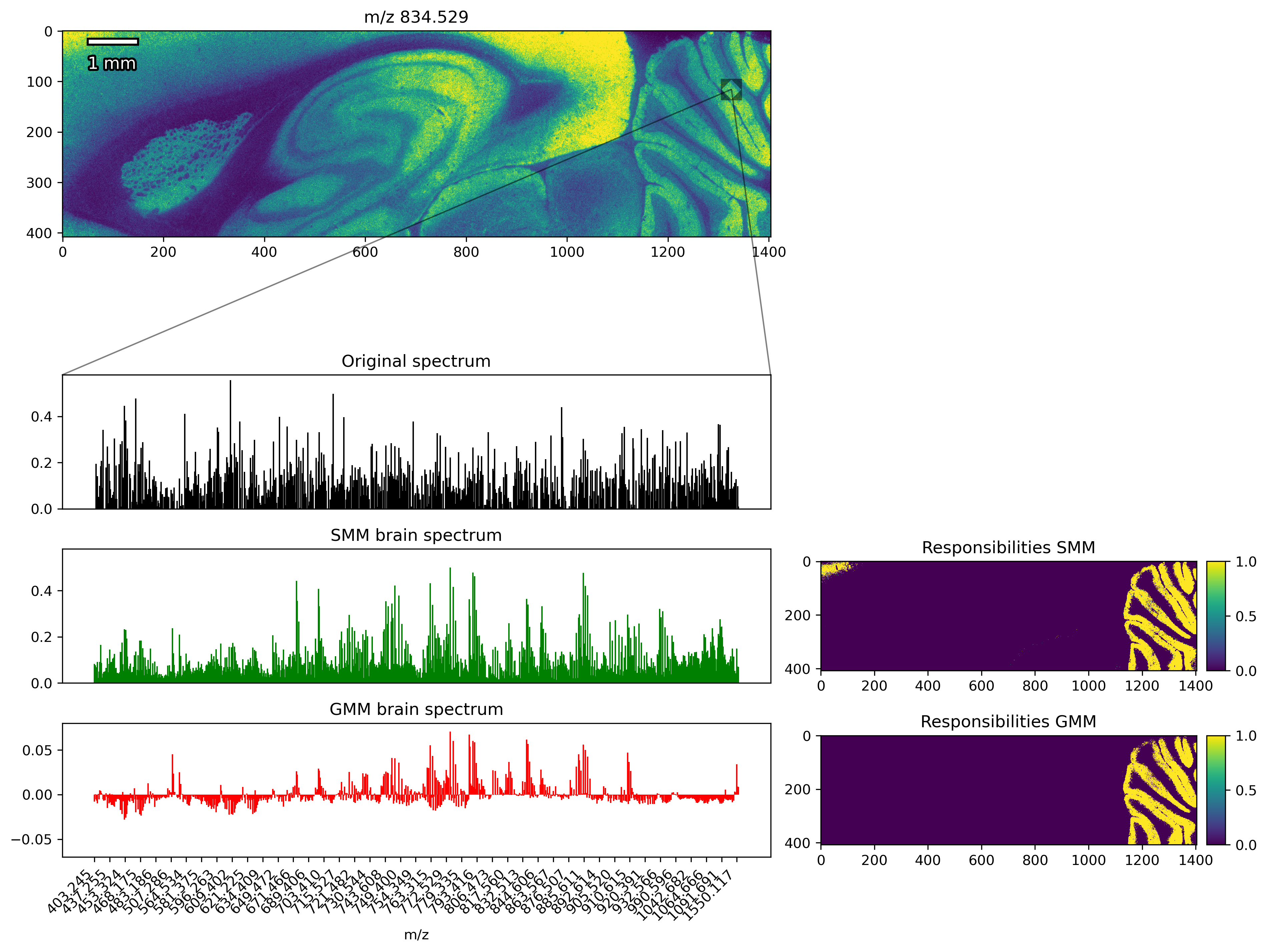}
    \caption{Comparison of \ac{SMM} and \ac{GMM} estimated subpopulation spectra for the rat brain \ac{IMS} dataset.
    (black trace) Original peak-picked mass spectrum acquired at the tissue location indicated above.
    (green trace) \ac{SMM}-estimated underlying signal for that same location.
    (red trace) \ac{GMM}-estimated underlying signal for that same location.
    Signals estimated by \ac{SMM} seem closer to real mass spectra than \ac{GMM}-estimated signals.
    For example, \ac{IMS} data is inherently non-negative.
    Without explicitly imposing non-negativity, \ac{SMM}-recovered signals are largely non-negative and more similar to real mass spectra than the \ac{GMM}-recovered signals, which exhibit substantial amounts of negative values.
}
    \label{fig:spectra_comparision_rat}
\end{figure*}

In Figure~\ref{fig:spectra_comparision_rat}, we provide a comparison between a noisy observation $\y_i$ and its underlying signal subpopulation $\x_i$ as estimated by \ac{EM}-based \ac{SMM} and \ac{GMM} fitting.
The black trace shows $\y_i$, an example peak-picked mass spectrum (before min-max normalization) measured at a particular location (pixel $i$) in the rat brain.
The red trace shows $\x_{i,\textrm{GMM}}$, the signal subpopulation estimated to be underlying observation $\y_i$ by fitting a \ac{GMM} (with the min-max normalization undone).
The green trace shows $\x_{i,\textrm{SMM}}$, the signal subpopulation or spike estimated to be underlying observation $\y_i$ by fitting a \ac{SMM} (with the min-max normalization undone).
We see that \ac{GMM} fails to capture the inherent non-negative nature of the data and produces a mix of positive and negative peaks in its recovered signals.
This is not the case for \ac{SMM}, which despite non-negativity not being imposed, recovers signals that are largely non-negative and that are much closer to measured mass spectra than the signals estimated by \ac{GMM}.
Additionally, \ac{SMM} seems to capture the low- and high-intensity patterns of the mass spectrum more accurately, particularly when considering the intensity ranges in which the recovered signals are provided.

\subsection{Rat brain \ac{IMS} dataset pre-processing}
\label{app:rat_brain_acquisition}

The \ac{IMS} dataset was exported from the Bruker timsToF file format (.d) to a custom binary format. 
Each pixel, \textit{i.e.}, `frame', reports between 10$^4$ and 10$^5$ centroided peaks that cover the entire acquisition range and that can be used to reconstruct a pseudo-profile mass spectrum using Bruker's SDK (v2.21) \cite{brukerSDK}.
The dataset was \ac{mz}-aligned by means of internally identified peaks (six peaks that appear in at least 50$\%$ of the pixels), using the msalign library (v0.2.0) \cite{monchamp2007signal,msalign}.
This step corrects for spectral misalignment (drift along the \ac{mz}-domain), resulting in increased correspondence and overlap between spectral features (peaks) across the experiment.
Subsequently, the mass axis of the dataset was calibrated, using theoretical masses for the same six peaks, to approximately $\pm$1 ppm precision.
After alignment and calibration, an average mass spectrum based on all pixels in the dataset was computed.
The average spectrum was peak-picked, 843 peaks were detected, and their corresponding ion intensities were retrieved for further analysis.
Isotopes were not removed and, instead, were allowed to take part in the downstream analysis.
Subsequently, we computed normalization correction factors using a total ion current (TIC) approach and intensity-normalized the ion counts to yield the \ac{IMS} dataset used in this paper.

\begin{figure*}[ht]
    \centering
    \includegraphics[width=0.9\textwidth]{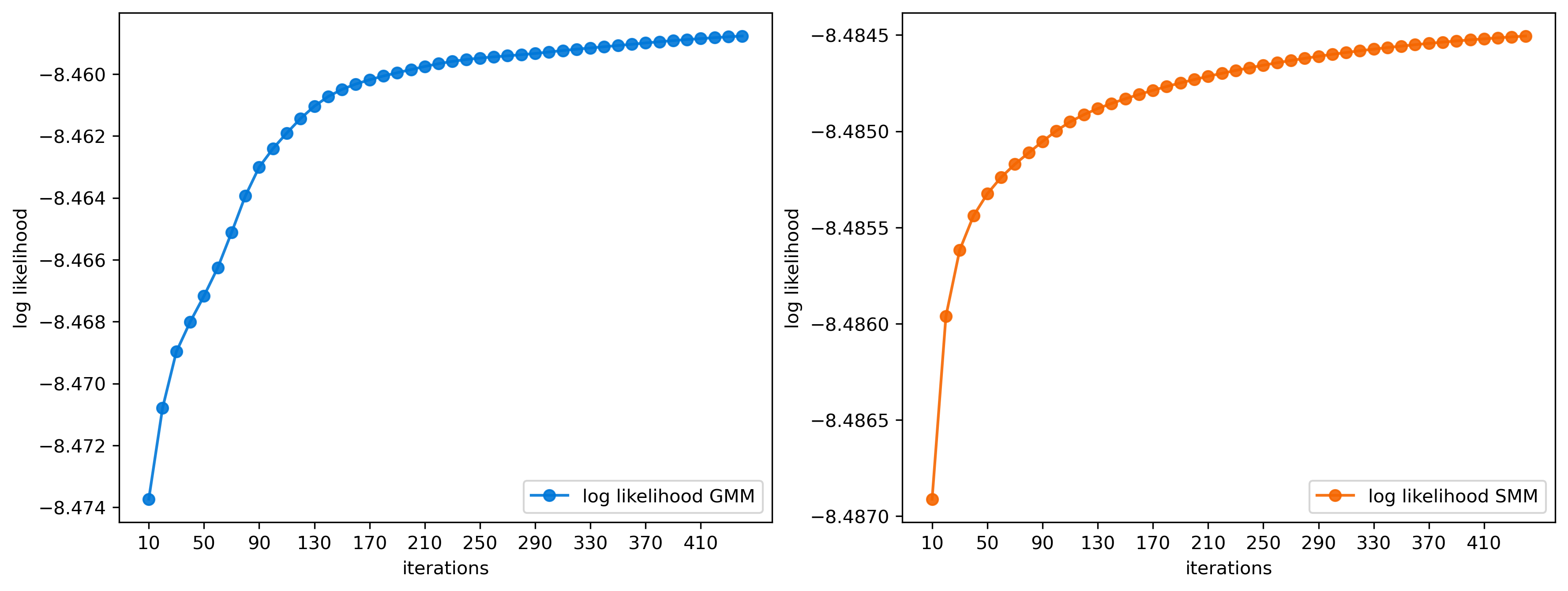}
    \caption{
    \textcolor{manualcolor}{
    Comparison of the normalized log-likelihood evolution for the \ac{GMM} (left) and \ac{SMM} (right) approaches, averaged over 100 repetitions, using an \ac{SMM} dataset with $N = 1500, d = 5, K=3$, and $\sigma^2 = 1.5$.
    }
    }
\label{fig:likelihood_evolution}
\end{figure*}

\textcolor{manualcolor}{
\subsection{Convergence speed comparison of likelihood estimation}
\label{sec:convergence_speed}
Although comparing the convergence speed of two different \ac{EM} algorithms is generally challenging, our goal here is to assess how the proposed \ac{SMM} approach performs relative to the traditional \ac{GMM} in terms of likelihood maximization.
To that end, we create a dataset of model~\eqref{eq:model} with $N = 1500, d = 5, K=3$, and $\sigma^2 = 1.5$. We repeat $100$ runs of the associated \ac{EM}-algorithm for both the \ac{SMM} and the \ac{GMM} and recall the average likelihood every $10$ iterations. The results are depicted in Figure~\ref{fig:likelihood_evolution}.
The shape of both likelihoods seems to suggest that \ac{SMM} converges relatively faster than \ac{GMM} to a local maximum. This could be explained by the number of modeled parameters that is smaller for the \ac{SMM}, leading to a smoother likelihood function.
}

\textcolor{manualcolor}{
\subsection{Convergence results of the \ac{EM} algorithm for model~\eqref{eq:model}
\label{sec:convergence_results_em}
}
We want to show that the likelihood value converges, and that if the parameters $\theta^{[t]}$ converge, then the limit will be a stationary point of the likelihood. These results hold with probability one if $N\ge d \ge K+1$.
\\
As mentioned in section~\ref{sec:expected_maximization}, improvements on the $\mathcal{Q}$ function guarantee improvements of the log-likelihood and thus of $L(\theta)$. This ensures that the sequence $\{L(\theta^{[t]})\}_t$ is not decreasing. \\
Looking at our $L(\theta)$ function described in \eqref{eq:p-of-y} and \eqref{eq:likelihood}, we see that $\sigma^2$ should be bounded from below to ensure $L(\theta)$ is bounded from above, which then implies convergence of the sequence $\{L(\theta^{[t]})\}_t$ \cite[section 3.4.1]{mclachlan2008algorithm}. Such a bound is shown to hold with probability one in Lemma~\ref{lem:sigma_lb} below, under the assumption that $N\geq d \geq K+1$.
\\
Next, we will use a general convergence result for EM algorithms which says that, if the parameter vector $\theta^{[t]}$ converges, then it is to a stationary point of the likelihood function $L(\theta) = p_\theta(y_1,\ldots,y_N)$, assuming that $L(\theta)$ is bounded and the function $\mathcal{Q}$ is continuous \cite[Chapter 3]{mclachlan2008algorithm}.
\\
\cite[Theorem 3.2]{mclachlan2008algorithm} guarantees that if $\mathcal{Q}(\theta;\theta^{[t]})$ is continuous in both $\theta$ and $\theta^{[t]}$, then all limit points of any instance $\{\theta^{[t]}\}$ of the \ac{EM} algorithm are stationary points of $L(\theta)$, and $\{L(\theta^{[t]})\}_t$ converges monotonically to some value $L^\star = L (\theta^\star)$ for some stationary point $\theta^\star$.
For a non-zero $\sigma$ (ensured by Lemma~\ref{lem:sigma_lb}), this continuity condition is fulfilled with probability one since the function $\mathcal{Q}(\theta;\theta^{[t]})$ given in \eqref{eq:Q_function_SMM} is a combination of continuous functions.
\\
\begin{lemma}
    For $d \geq K+1$, at every iteration $t$, we have
    \begin{align} \label{eq:lemma-C3-bound}
        \sigma^{2,[t]} \geq \frac{\lambda_{K+1}(\Y\Y^T)}{dN}.
    \end{align}
    Furthermore, for observations coming from the model~\eqref{eq:model}, 
    \begin{align*}
        \mathbb{P}\left[\frac{\lambda_{K+1}(\Y\Y^T)}{dN} > 0\right] = 1
    \end{align*}
    whenever $N \geq d$.
    \label{lem:sigma_lb}
\end{lemma}
\begin{proof}
    For $S \subseteq [K]$, the expression of the noise variance at every step of the \ac{EM} algorithm is
    \begin{align*}
        \sigma^2 &= \sigma^2(S):= \frac{ \|\Y\|_\mathrm{F}^2 -\sum_{k \in S} \lambda_k}{dN - \sum_{k \in S}\gamma_k }, &\textrm{from } \eqref{eq:sigma}
    \end{align*}
    where $\lambda_k$ is the largest eigenvalue of $\A_k = \sum_{i=1}^N \rho_k^{[i]} \y_i \y_i^T$.
    \\
    Similar to the proof of Lemma~\eqref{lem:sigma_positive}, we find
    \begin{align}
        \|\Y\|_F^2 - \sum_{k \in S} \lambda_k &\geq \|\Y\|_F^2 - \sum_{k=1}^K \lambda_k &S \subseteq [K] \notag\\
        &= \text{Tr}\left(\Y\Y^T\right) - \sum_{k=1}^K \lambda_1(\A_k) \notag \\
        &=\sum_{k=1}^K \text{Tr}\left(\A_k\right)-\lambda_1(\A_k),  &\sum_{k=1}^K \A_k = \Y\Y^T \notag \\
        &= \sum_{k=1}^K \sum_{j=2}^d \lambda_j(\A_k) \notag \\
        &\geq \sum_{k=1}^K \lambda_2(\A_k).
        \label{eq:numerator_sigma2}
    \end{align}
    Furthermore, since $0 \leq \sum_{k\in S}\gamma_k \leq N$, we have 
    \begin{align}
        0 \leq dN - \sum_{k\in S} \gamma_k \leq dN.    
        \label{eq:denominator_sigma2}
    \end{align}
    Equation \eqref{eq:numerator_sigma2} together with \eqref{eq:denominator_sigma2} leads to 
    \begin{align*}
        \sigma^2 \geq \frac{\sum_{k=1}^K \lambda_2(\A_k)}{dN}.
    \end{align*}
    From here, the first part of the lemma (\textit{i.e.}, eq.~\ref{eq:lemma-C3-bound}) follows with the following lower bound on $\sum_{k=1}^K \lambda_2(\A_k)$ depending only on the matrix $\Y$.
    Using Weyl's inequality recursively, 
    \begin{align*}
        \lambda_{K+1}\left(\Y\Y^T\right) &= \lambda_{K+1}\left(\sum_{k=1}^K \A_k\right) \\
        &\leq \lambda_2(\A_1) + \lambda_K\left(\sum_{k=2}^K\A_k\right) \\
        &\leq \lambda_2(\A_1)+ \lambda_2(\A_2)+ \lambda_{K-1}\left(\sum_{k=3}^K\A_k\right) \\
        &\leq \sum_{k=1}^K \lambda_2(\A_k).
    \end{align*}
    The second part of the lemma follows by showing that the right-hand side of \ref{eq:lemma-C3-bound} is almost surely strictly positive.
    This is surely the case when $\Y\Y^T$ has full rank.
    The set of vectors $y_1, \ldots, y_N$ that lie in a lower-dimensional subspace of $\mathbb{R}^d$ has zero Lebesgue measure. Since the joint distribution of $\y_1,\ldots, \y_N$ is absolutely continuous with respect to the Lebesgue measure, the matrix $\Y$ is therefore of rank $d$ with probability $1$. For $N\geq d$, this ensures that $\lambda_{K+1}(\Y\Y^T) > 0$ almost surely.
\end{proof}
}

\vfill

\end{document}